\documentclass[journal]{IEEEtran}
\usepackage{times}
\usepackage{algorithm}
\usepackage{algpseudocode}
\usepackage{helvet}
\usepackage{courier}
\usepackage{url}
\usepackage{graphicx}
\usepackage{amssymb}
\usepackage{subfig}
\usepackage{amsmath}
\usepackage{booktabs}
\usepackage{tabularx}
\usepackage{multirow}

\begin{document}	

\title{Network Embedding via Deep Prediction Model\\}
%
%
%
\author{Xin Sun,~\IEEEmembership{Member,~IEEE,}
        Zenghui Song, Yongbo Yu, 
        Junyu Dong, ~\IEEEmembership{Member,~IEEE,},
        Claudia Plant, ~\IEEEmembership{Member,~IEEE,},
        and Christian B\"{o}hm,~\IEEEmembership{Member,~IEEE}
\thanks{X. Sun, Z. Song, Y. Yu, J. Dong are with Department of Computer Science and Technology, Ocean University of China, Qingdao, China. X Sun is also with the Department of Aerospace and Geodesy, Technical University of Munich, Germany.}
\thanks{C. Plant is with Faculty of Computer Science, University of Vienna, Vienna, Austria and Data Science @ University of Vienna, Vienna, Austria.}
\thanks{C. B\"{o}hm is with Ludwig-Maximilians-Universit\"{a}t M\"{u}nchen, Munich, Germany}}

%
%

\markboth{Journal of \LaTeX\ Class Files}%
{Shell \MakeLowercase{\textit{et al.}}: Bare Demo of IEEEtran.cls for IEEE Journals}
%



\maketitle

\begin{abstract}
Network-structured data becomes ubiquitous in daily life and is growing at a rapid pace. It presents great challenges to feature engineering due to the high non-linearity and sparsity of the data. The local and global structure of the real-world networks can be reflected by dynamical transfer behaviors among nodes. This paper proposes a network embedding framework to capture the transfer behaviors on  structured networks via deep prediction models. We first design a degree-weight biased random walk model to capture the transfer behaviors on the network. Then a deep network embedding method is introduced to preserve the transfer possibilities among the nodes. A network structure embedding layer is added into conventional deep prediction models, including Long Short-Term Memory Network and Recurrent Neural Network, to utilize the sequence prediction ability. To keep the local network neighborhood, we further perform a Laplacian supervised space optimization on the embedding feature representations. Experimental studies are conducted on various datasets including social networks, citation networks, biomedical network, collaboration network and language network. The results show that the learned representations can be effectively used as features in a variety of tasks, such as clustering, visualization, classification, reconstruction and link prediction, and achieve promising performance compared with state-of-the-arts.
\end{abstract}


%
\IEEEpeerreviewmaketitle

\section{Introduction}
%
%
%
%
\IEEEPARstart{N}{owadays} structured data in the form of networks is ubiquitous in
our daily lives and has an astonishing growth \cite{Zhang2017Network,Hamilton2017}.
Especially the pervasive use of online social networks, such as Facebook and Twitter, generates huge network data continuously at unprecedented rates. New applications, such as node classification \cite{Sen2008,Zhu2018}, link prediction \cite{Liben2007,Li8450054}, social role discovering \cite{Henderson2012,Tu2018}, community detection \cite{Cheng8531771,Li2018} and recommendation \cite{Ying2018,Cai2018}, arise in various areas. For example, node classification and link prediction are commonly used for similar user searching and advertisement recommendation \cite{Cai2017,Goyal2018}. However, the vast majority of existing machine learning algorithms for classification and prediction are feature-based \cite{Wold1987}, which means informative and discriminating attribute-value entities are required. Rather than developing special learning algorithms for network data, it is more practical to learn feature vector representations for nodes and edges of the network. Once the vectorized representation is obtained, the data mining tasks for large networks can be well solved by state-of-the-art machine learning algorithms. Therefore, the issue of how to transfer the link-connected nodes of the huge network into feature representations is critical.
Intuitively the widely studied graph embedding approaches including IsoMap \cite{Tenenbaum2000}, LLE \cite{Roweis2000}, and Laplacian Eigenmaps \cite{Belkin2001} seem to be good solutions for the network feature learning problem. Nevertheless, the graph representation, which is the core of these approaches, is directly derived from the data itself and perfectly reflects the global and local structure of the data. On the contrary, natural networks encountered in the real world are sparse and many undiscovered links are noisy \cite{Tang2012}. We cannot get a full view of the authentic relationship between any two nodes. So it is impractical to get features from the network itself by traditional graph embedding approaches. One typical solution is to form hand-engineering features for each node by observing its interaction behavior with neighbors \cite{Perozzi2014}, such as first-order proximity and second-order proximity \cite{Tang2015}. However, the neighbor context information is still not enough to well describe the node due to the undiscovered and noisy links. The networks from the real world exhibit significant dynamic evolution behavior \cite{Li2017}. It is impossible to catch the network evolution process from the structured data itself. Fortunately, the dynamic evolution process can be reflected by the transfer behaviors among the nodes, such as rumor propagation and reputation transmission.

Previous work did not systematically consider local and global structure with its dynamical behaviors, so this paper proposes a Network Embedding framework via Deep Prediction models to solve this issue in two aspects. One is how to capture the network structure and neighborhood context information from the network structured data. The other is how to embed the complex and non-linear network structural data into low-dimensional vector representations, while preserving the transfer possibilities among the nodes.
Our contributions are as follows:
\begin{itemize}
	\item We propose a degree-weight biased random walk model to preserve the global network context information. It captures the node's roles and the information transfer behaviors among nodes.
	\item To preserve the transfer possibilities among the nodes, we design a deep network embedding framework, which suggests a network structure embedding layer into conventional deep prediction models, such as Long Short-Term Memory Network (LSTM) and Recurrent Neural Network (RNN), to utilize their sequence prediction ability.
	\item We leverage a Laplacian supervised embedding space optimization to capture the local network structure, which makes the connected nodes close with each other in the low-dimensional space.
	\item We conduct extensive experiments on various tasks including such as clustering, visualization, classification, reconstruction and link prediction, and achieve competitive performance.
\end{itemize}

The rest of this paper is organized as follows. Section II summarizes the related works. Section III formally gives some problem definitions for network data. Section IV introduces our model in detail. Section V presents the experimental results. Finally, we conclude in Section VI.

\section{Related Work}
This work focuses on engineering features from the network structured data. A practical way to engineer features for the network structured data is to design domain-specific features based on expert knowledge. For example, Henderson et al. \cite{Henderson2011} introduced the statistical properties of the node itself (e.g., degree) and its neighborhood (number of edges) to generate the recursive structural features. With their structural features, a role discovery approach RolX \cite{Henderson2012} was designed to mine similar nodes with similar social behavior. Inspired from the representational learning for natural language processing, such as Skip-gram model \cite{Mikolov2013}, Perozzi et al. \cite{Perozzi2014} proposed the DeepWalk method by treating nodes in the network as words. Their motivation is that the distributions of vertices in social network and words in natural language appearing in short random walks both follow a power-law behavior. Similarly, node2vec \cite{Grover2016} offers flexibility random walk sampling strategies to capture the diversity of the neighborhood. However, these methods do not make clear what kind of network properties are preserved. LINE \cite{Tang2015} method defines the first-order and second-order proximities to clearly preserve both the local and global network structures. AROPE \cite{Zhang2018ArbitraryOrderPP} abandons absolute proximity while it uses SVD framework to learn network representations from the perspective of the proximity of arbitrary order. HOPE \cite{Ou2016} solves the asymmetric transitivity problem in a directed network and preserves high-order proximities of large-scale graphs with generalized SVD. NetHiex \cite{Ma2018HierarchicalTA} incorporates the hierarchical taxonomy into network representation learning, which can capture the latent hierarchical taxonomy. 

The network data is highly non-linear \cite{Tang2012,Zhang8395024} with various links, Tian et al. \cite{Tian2014} proposed a deep autoencoder to learn nodes' representation which was used for graph clustering. And Wang et al. \cite{Wang2016} designed a semi-supervised deep model to capture the highly non-linear network structure based on the first-order and second-order proximity. DNGR \cite{Cao2016} uses the stacked denoising autoencoder to learn the low-dimensional node representations. ANRL \cite{Zhang2018} utilizes one autoencoder model and one skip-gram model to learn the representations of the different node jointly. Gao et al. \cite{Gao2018} used two parallel autoencoders to obtain network representation respectively, then concatenate the two representation as a result. Cavallari et al. \cite{Cavallari2017LearningCE} and Wang et al. \cite{Wang2017CommunityPN} both proposed to learn network representation by community. The former designed a framework to unify the node embedding, community embedding, and community detection together to get robust representations. And the latter exploited the consensus relationship between the representations of nodes and community structure, and jointly optimized NMF based representation learning model and modularity based community detection model in a unified framework. 

As the convolutional neural network (CNN) has achieved great success in the image processing area \cite{Krizhevsky2012}, some works were carried out with CNN \cite{Bruna2013}. Inspired by spectral convolutions, Kipf et al. \cite{Kipf2016} proposed an efficient convolutional neural network, named GCN  that can classify the network data. FastGCN \cite{Chen2018} improved the efficiency for GCN in graph neural network by importance sampling. Rahimi et al. \cite{Rahimi2018SemisupervisedUG} took advantage of the GCN to do multiview user geolocation. Furthermore, some work focus on transforming the structure-like data into grid-like data such as images in some ways, then the traditional convolutional operator can be applied to transformed data. Niepert et al. \cite{Niepert2016} tried to construct a convolution operator from spatial domain instead of the spectral domain for the network data. Gao et al. \cite{Gao2018} proposed a graph convolution layer to transform the structure data to grid data, which applied the convolutional operator naturally. Shi et al. \cite{Shi9177263} proposed a multi-label graph convolutional network (MuLGCN) for learning node representation of multi-label networks. 

Besides the CNN deep model, Generative Adversarial Networks (GAN) \cite{Goodfellow2014} is also introduced to handle the network structured data. For instance, Pan et al. \cite{Pan2018}, Dai et al. \cite{Dai2017} and Yu et al. \cite{Yu2018} all used the GAN model to learn the network representation. The key to use GAN model is how to choose the positive sample and the negative sample. The three works mentioned above used different positive sample and negative sample. Dai et al. \cite{Dai2017} and Pan et al. \cite{Pan2018} both used the prior distribution(e.g. uniform distribution or normal distribution) as the positive sample, while Yu et al. \cite{Yu2018} chose the generated representation as the positive.

So far, some of the network embedding research works were already applied to specific tasks \cite{Zhang8395024,Song9241052}. For example, PME \cite{Chen2018PMEPM} captures both first-order and second-order proximities in a unified way based on metric learning in the link prediction task. SEMAC \cite{Cao2018LinkPV} exploits fine-grained node features as well as the overall graph topology to learn representations. Especially, SHINE \cite{Wang2018SHINESH} works on sentiment link prediction where they utilized multiple deep autoencoders to map each user into a low dimension feature space while preserving the network structure. In classification tasks, Nandanwar et al. \cite{Nandanwar2016StructuralNB} proposed a structural neighborhood-based classifier learning using a random walk. DDRW \cite{Li2016DiscriminativeDR} improves the DeepWalk by solving a joint optimization problem to learn the latent space representations that captured the topological structure. Jacob et al. \cite{Jacob2014LearningLR} aimed to learn the heterogeneous network's representation by a new framework in which two nodes connected will tend to share similar representations regardless of their types. In name disambiguation task, Zhang et al. \cite{Zhang2017NameDI} utilized network representation learning on the relational graph data to embed nodes in low dimensional space. In clustering tasks, MCGE \cite{Ma2017MultiviewCW} models the graph as tensors and applied tensor factorization to learn the graph embeddings. In anomaly detection tasks, NetWalk \cite{Yu2018NetWalkAF} first encodes the nodes to feature representations by clique embedding, then it jointly minimizes the pairwise distance of node representations. DynamicTriad \cite{Zhou2018DynamicNE} learns the representations by modeling  closed triad which consists of three nodes connected with each other. Cai et al. \cite{Cai2018GenerativeAN} proposed a deep network representation model that integrated network structure and the node content information by GAN, which can be applied to personalized citation recommendation. 

Although the above methods have achieved good performance on some tasks, the structure of the network they considered are not comprehensive and the way of choosing positive and negative sample is confusing.

This section cannot be a complete review of all algorithms and just briefly describes the most related studies.

\begin{figure*}
	\centering
	\includegraphics[width=1.01\textwidth]{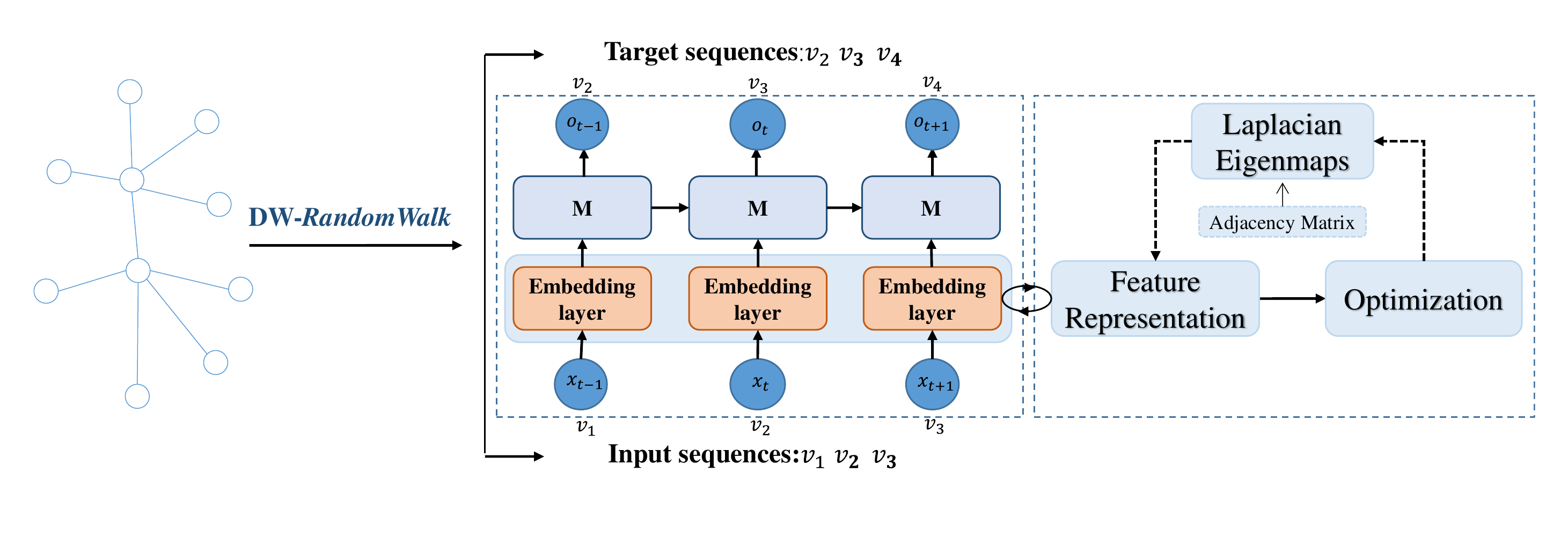}
	\caption{The proposed Network Embedding framework via Deep Prediction models (NEDP) framework. $M$ is an available modular which includes RNN and LSTM.}
	\label{fig:framework}
\end{figure*}

\section{Problem Definition}
In this section, we define the feature engineering problem for network data and give some concepts of network structured data concerned in this paper.

\textbf{Definition 1}. The network structured data can be represented as $N$=\{$V$, $E$\}. $V$=\{$v_1$, $v_2$, $\cdots$, $v_n$\} is the node set where $n$ is the number of nodes. $E=\{e_{ij}\}_{i,j=1}^n$ is the edge set where $e_{ij}$ denotes the edges from node $v_i$ to $v_j$. A weight $w_{ij}\geq0$ is attached to each edge $e_{ij}$, and $w_{ij}=0$ if $v_i$ and $v_j$ are not directly connected. For unweighted network $w_{ij}=1$, and for undirected network $w_{ij}=w_{ji}$.

\textbf{Definition 2}. (\textit{Network numerical formalization}) The network feature representation learning problem aims to find a robust and low-dimensional vector representation for each node. This paper addresses this problem in two steps: (1) represent the structural network data in the numerical vector space that capture the structural function and global neighborhood context information; (2) encode the complex and non-linear network structural representations into low-dimensional vector representations, while preserving the transfer possibilities among the nodes. Then this problem can be defined as follows.

Given a network $N$=\{$V$, $E$\}, numerical formalization aims to represent each vertex $v_i\in V$ as a $D$-dimensional numerical vector $x_i\in \mathbb{R}^D$. The  numerical vector $x_i$ preserves the neighborhood and structural role information.
(\textit{Feature learning}) It aims to learn a function $f$: $x_i\rightarrow y_i\in\mathbb{R}^d$, where $d\ll D$. The objective of the function $f$ is to minimize the distance between $y_i$ and $y_j$ if they are similar to each other.

In the next section, we will introduce the network numerical formalization and feature learning respectively.

\section{Model Description}

In this section, we introduce a Network Embedding framework via Deep Prediction models (NEDP) for representation learning as shown in Figure \ref{fig:framework}. The NEDP framework consists of three parts: (i) Degree-Weight biased random walk (DW-$RandomWalk$) for sampling sequences, (ii) A network structure embedding layer of the deep prediction model for encoding high-dimensional space into low-dimensional one, (iii) Laplacian \cite{Belkin2001} supervised embedding space optimization (LapEO) for capturing the local network structure.

\subsection{Degree-Weight biased Random Walk}
Network structure data is highly complex and non-linearly, so either it is almost impossible to calculate these data directly with a deep learning model, or the calculation cost is particularly expensive even though choose some special model, like autoencoder. To overcome the above problem, we apply random walk algorithm, which is used in DeepWalk \cite{Perozzi2014} and Node2Vec \cite{Grover2016}, to generate a large number of node sequences for training.
DeepWalk used the truncated random walk to generate node sequences which were feed to skip-gram model to obtain network representation. Given a source node $u$, we get a sequence of length $l$ by truncated random walk \cite{Fouss2007}. The sequences starting from node $u$ are generated by the following probability:
\begin{equation}\label{eq:trw}
	P_t(u\to{x})=\frac{1}{\left|{\mathcal{N}(u)}\right|},x\in{\mathcal{N}(u)}
\end{equation}

where $P_t$ denotes the transition probability of truncated random walk\cite{Perozzi2014}, $\mathcal{N}(u)$ denotes the node $u$ neighbors. However, it will miss many rich neighbors information because truncated random walk only considered the number of current node's neighbors.
To overcome the challenge, Node2Vec proposed biased random walk which set a search bias $\alpha$ to guide the walk, so we set the unnormalized transition probability to
\begin{equation}\label{eq:brw}
	P_b(u\to{x})=\alpha * \omega_{ux},x\in{\mathcal{N}(u)},\alpha\in{\{\frac{1}{q},\frac{1}{p},1\}}
\end{equation}

where $P_b$ denotes the unnormalized transition probability of biased random walk, $\omega_{ux}$ denotes the edge $(u,x)$ weights, and $\alpha$ will be chosen according to the  distance between two nodes, $p$ and $q$ are hyper-parameters\cite{Grover2016}. Through introducing the above parameters, the biased random walk can automatically choose the Depth-first Sampling or Breadth-first Sampling to generate node sequences including diverse neighbors.

\begin{figure}
	\centering
	\includegraphics[width=0.35\textwidth]{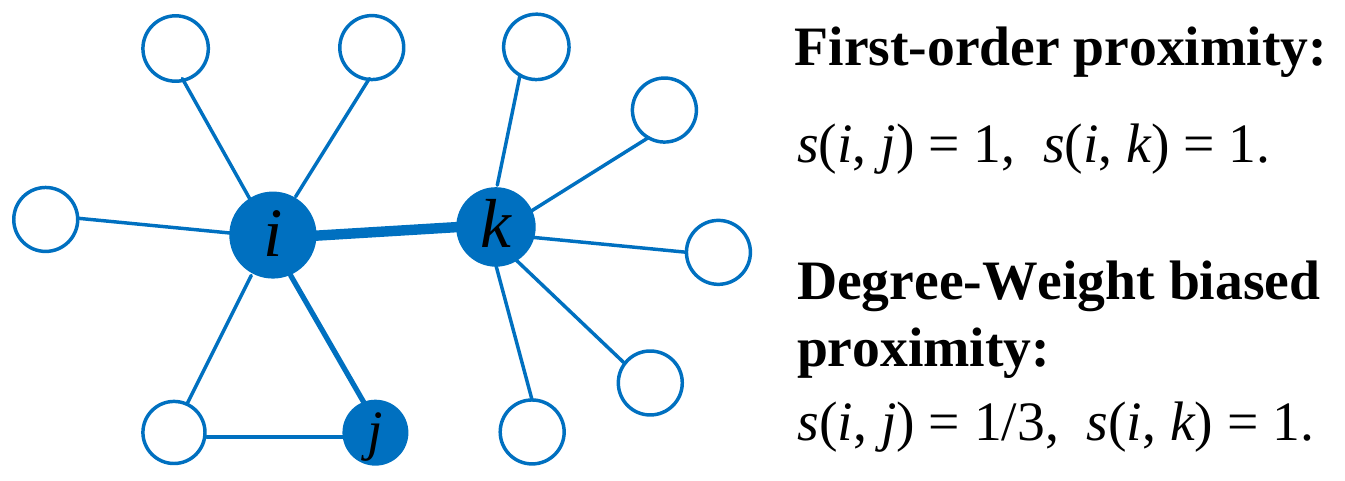}
	\caption{A toy example of subgraphs. Social role and relationship is quite different among  nodes $i$, $j$ and $k$.}
	\label{fig:example2}
\end{figure}

In most of the real networks, edge usually implies the similarity of two nodes, such as friend relationships. Some pieces of literature \cite{Wang2016} define the first-order proximity to characterize the pairwise proximity. The drawback is that the social role and relationship information are missing. As shown in Figure \ref{fig:example2}, the social role of nodes $i$ and $k$ is much more similar than $i$ and $j$ in term of function and structure. The information transfer more possibly exists between nodes $i$ and $k$. Here we propose a degree-weight biased proximity to characterize the transfer priority. For each pair of nodes connected by an edge $e_{ij}$, the proximity $s_{ij}$ is calculated as follow.
\begin{equation}
\label{equ:s_1}
s_{ij}=w_{ij}\cdot\frac{\min(d_i,d_j)}{\max(d_i,d_j)+ \alpha}
\end{equation}
The degree-weight biased proximity has the ability to capture the social role information among nodes. For example, we can see that proximity $s_{ik}$ and $s_{ij}$ in Figure \ref{fig:example2} is characterized correctly. That means nodes with similar function and role have high proximity probability. The $\alpha$ is an adjustment parameter which makes the $s_{ij}$ smooth.
We use degree-weight biased proximity $s_{ij}$ to guide the walking process. It tends to capture the primary structure characteristics among the nodes. The unnormalized transition probability is defined as follows:

\begin{equation}\label{equ:pd}
P(u^i\to{u^{i+1}})=s_{u^i, u^{i+1}},  u^{i+1}\in{\mathcal{N}(u^i)}
\end{equation}
where $u^i$, $u^{(i+1)}$ denote two connected nodes in networks. And the unnormalized transition probabilities of each node are normalized in order to make the summation equaling to 1. We call this sampling method as Degree-Weight biased Random Walk. Nowadays, there are many excellent sampling methods on network representation learning task, e.g., truncated random walk \cite{Perozzi2014} and biased random walk \cite{Grover2016}. We will demonstrate in the later section that our method by applying it to various real-world datasets is robust than other well-known methods.

\subsection{The Embedding Model}
In this section, we will introduce our core embedding model which is deep prediction network including Recurrent Neural Network and Long Short-Term Memory Network.
\subsubsection{Recurrent Neural Network}
\begin{figure}[ht]
\centering
\includegraphics[scale=0.47]{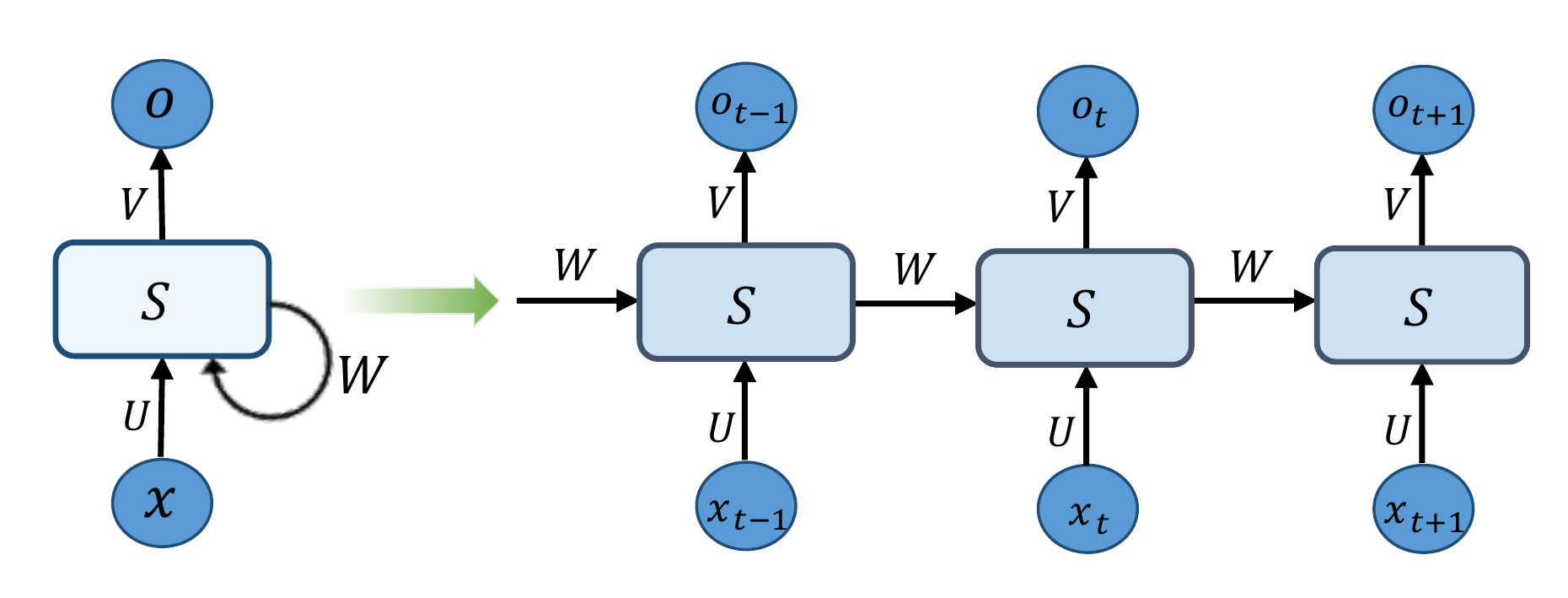}
\caption{Recurrent neural network (RNN)}
\label{fig:rnn}
\end{figure}

Recurrent Neural Network (RNN) was originally proposed to solve sequence problems in natural language processing. It can effectively predict the sequence of future time periods based on the given sequence. RNN is different from the traditional neural network (such as convolutional neural network and multiple layer perceptron) in the construction of the hidden layer. Figure \ref{fig:rnn} shows a simple RNN architecture which includes input layer, hidden layer and output layer.

It is obvious to observe that the state $s$ of the hidden layer of the RNN is affected by both the current input $x$ and the state value $s$ of the previous time step. $U$, $V$, $W$ respectively represent a weight matrix between the input layer and the hidden layer, a weight matrix between the hidden layer and the output layer, and a weight matrix between the hidden layers in Figure \ref{fig:rnn}. The forward propagation formula of RNN can be represented as follows:
\begin{equation}\label{eq:s_t}
	s_t = f(U\cdot x_t + W\cdot s_{t-1})
\end{equation}

\begin{equation}\label{o_t}
	o_t = g(V\cdot s_t)
\end{equation}

The above two formulas calculate the hidden and output layers respectively. $f(\cdot)$ and $g(\cdot)$ are both activation function. Generally, the activation function $f(\cdot)$ is $tanh$ which compresses the output of hidden layer to $(-1, 1)$. Under most circumstances, we choose $softmax$ as $g(\cdot)$ activation function because it will generate a normalized probability distribution which is usually applied to classification or prediction tasks. Actually, it is the data flow's recurrent mechanism in RNN's hidden layer that differs from CNN and MLP. It is worth noting that the weight matrix in RNN is shared.

\subsubsection{Long Short-Term Memory Network}
 \begin{figure}[ht]
 	\centering
 	\includegraphics[scale=0.195]{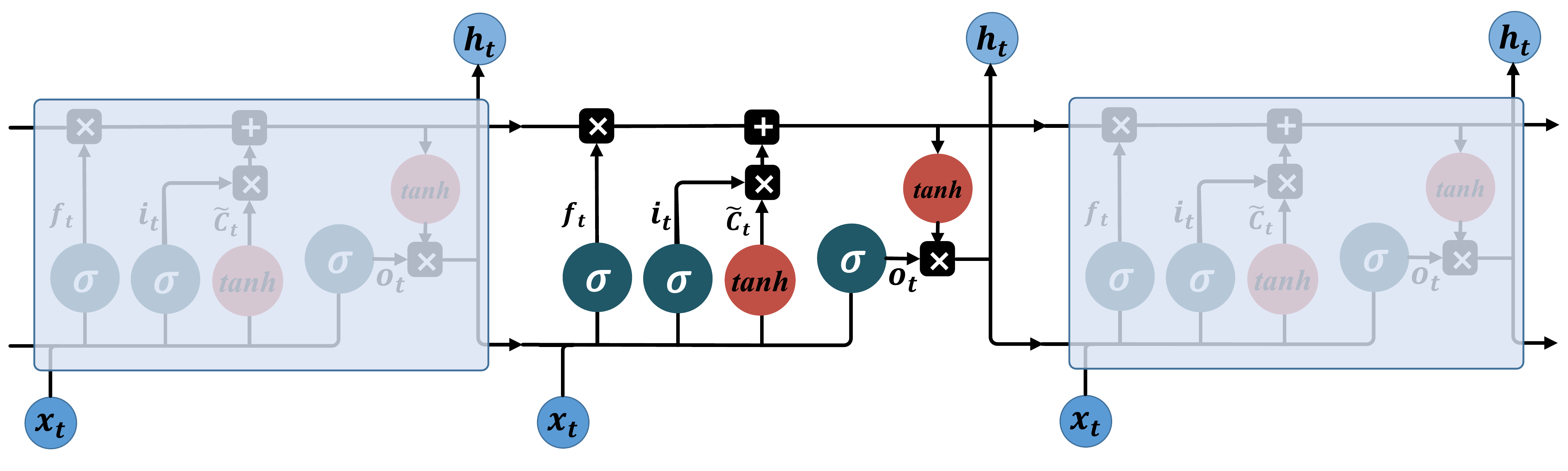}
 	\caption{Long Short-Term Memory Network (LSTM)}
 	\label{fig:lstm}
 \end{figure}
 \begin{figure*}[t]
 	\centering
 	\subfloat[LSTM forget gate]{\includegraphics[width=0.25\textwidth]{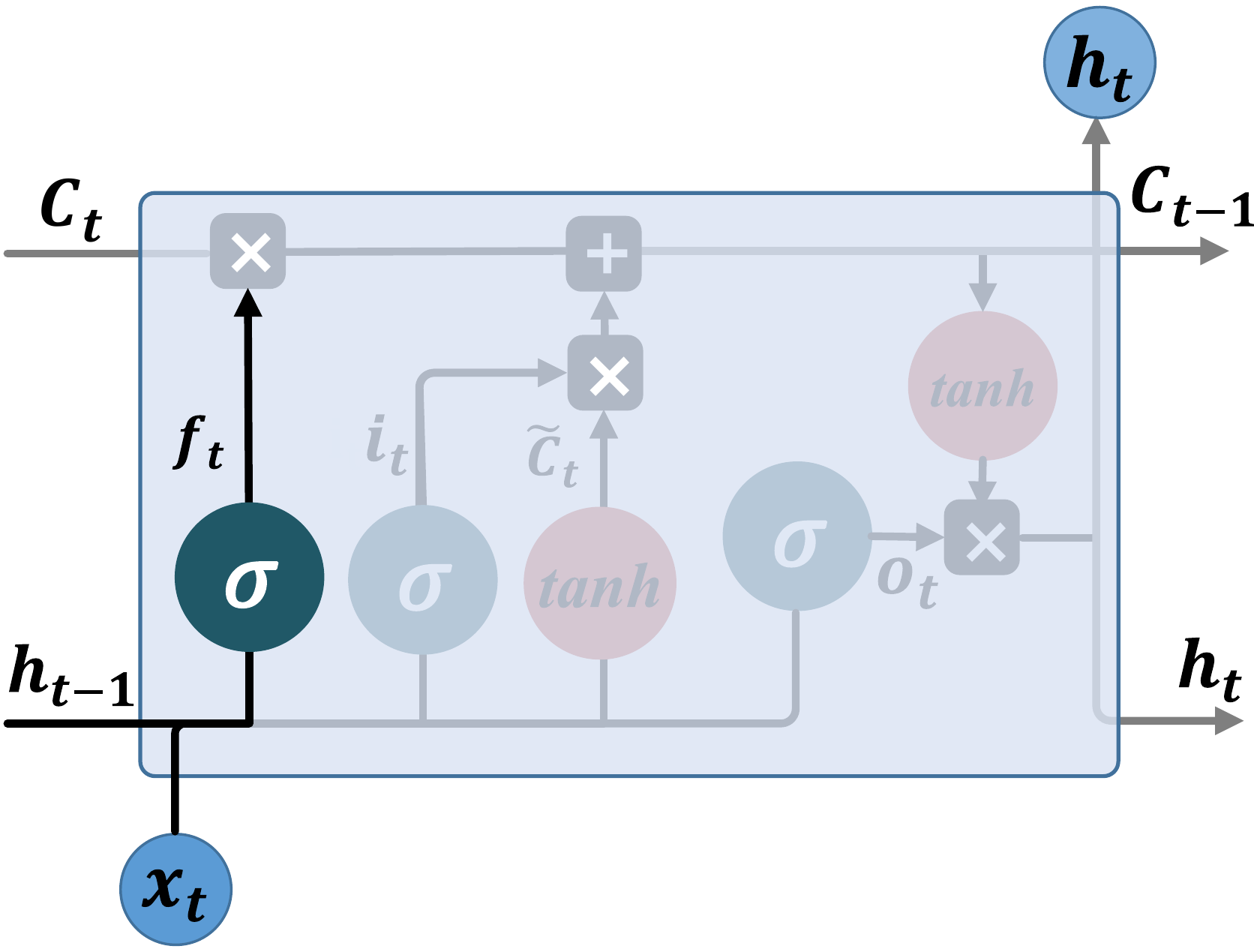}\label{fig:forget}}
 	\subfloat[LSTM input gate]{\includegraphics[width=0.25\textwidth]{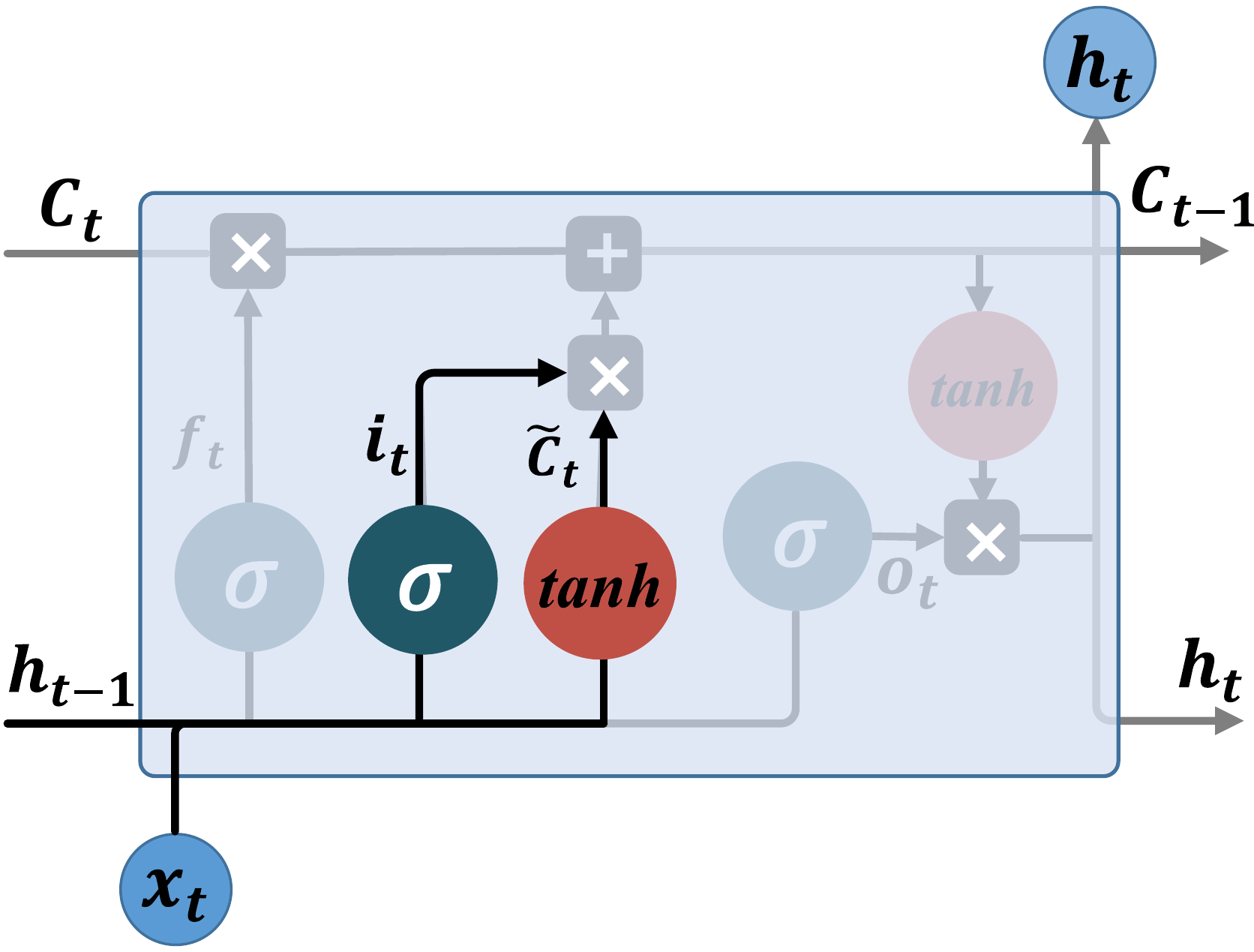}\label{fig:input}}
 	\subfloat[LSTM update]{\includegraphics[width=0.25\textwidth]{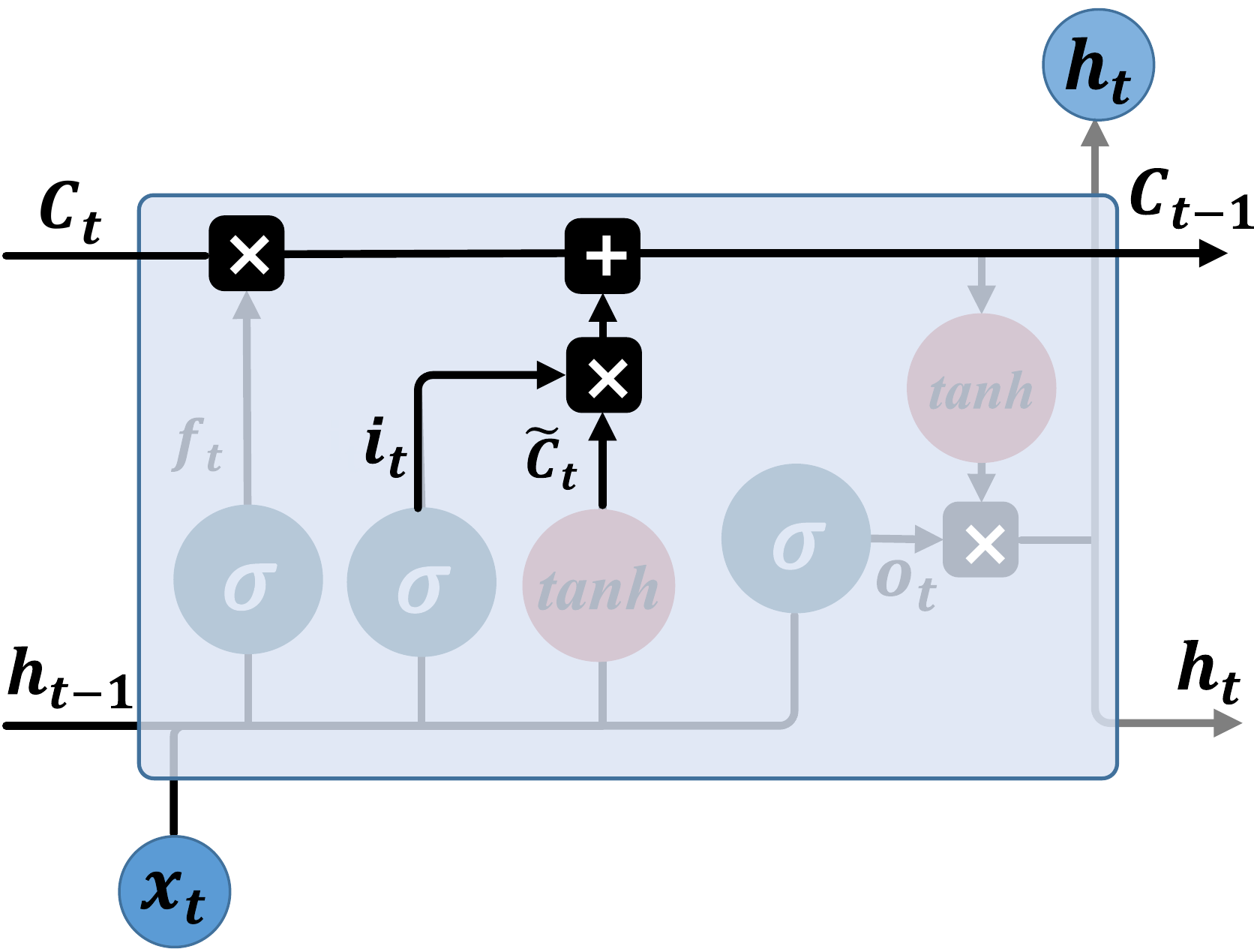}\label{fig:update}}
 	\subfloat[LSTM output gate]{\includegraphics[width=0.25\textwidth]{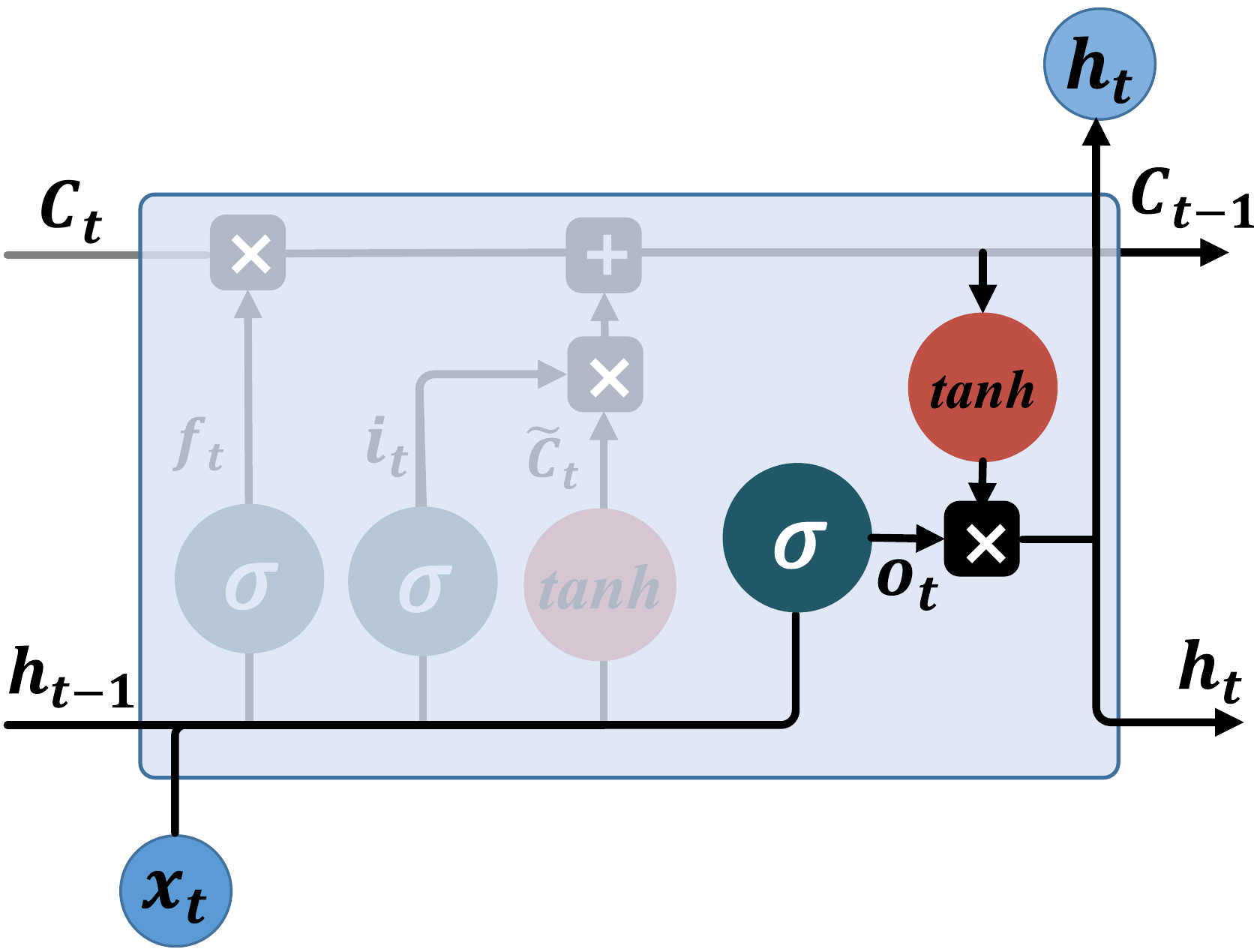}\label{fig:output}}
 	\caption{The core component in LSTM}
 	\label{fig:lstm_com}
 \end{figure*}
 
Since the RNN has the problem of the gradient vanishing, it cannot solve the problem of long-term dependence. To deal with the long-term dependence, Schmidhuber et al. \cite{Hochreiter1997} proposed Long Short-Term Memory Network (LSTM). RNN unit only has one state value $s$ which is very sensitive to short-term input. LSTM unit preserves long-term state by adding a new state $c$, called cell state. It is suitable for the speech recognition problem with huge vocabulary of words. Figure \ref{fig:lstm} shows the architecture of LSTM.

As shown in Figure \ref{fig:lstm}, the important part in LSTM is the cell state $C_t$. It is used to record the current information, which is obtained after the cell state of the previous moment was added and deleted, and pass them to the next block. LSTM controls the update of cell state by designing the structure of gate-input gate $i_t$ (Figure \ref{fig:input}), forget gate $f_t$ (Figure \ref{fig:forget}) and output gate $o_t$ (Figure \ref{fig:output}). LSTM units work by following ways.
At each time step, it takes into inputs from two external sources. One is the current input $x_t$, and one is the previous hidden state $h_{t-1}$. In addition, each gate has an internal input, the cell state $C_{t-1}$ of the previous LSTM units. The two external inputs first pass through a forget gate to determine which information to discard, then update the cell state. Figure \ref{fig:forget} shows forget gate and the formula is as follows:
 \begin{equation}\label{eq:f_t}
 	f_t=\sigma(W_f\cdot [h_{t-1},x_t] + b_f)
 \end{equation}

Next, we need to determine the new information added to the cell state. As show in \ref{fig:input}. The two external inputs go through an input gate layer and $tanh$ layer, which we will get the updated value through the input gate, and get a candidate state $\tilde{C_t}$ through the $tanh$ layer:
\begin{equation}\label{eq:i_t}
	i_t=\sigma(W_i\cdot [h_{t-1},x_t]+b_i)
\end{equation}

\begin{equation}\label{eq:cc_t}
	\tilde{C_t}=tanh(W_c\cdot [h_{t-1},x_t]+b_c)
\end{equation}

After the first two steps we got the cell state of the candidate cells $\tilde{C_t}$, the information $f_t$ we need to forget, the updated values $i_t$. As shown in \ref{fig:input}, we only need to update them to the cell state $C_t$ by the following formula:
\begin{equation}\label{eq:c_t}
	C_t = f_t \cdot C_{t-1} + i_t \cdot \tilde{C_t}
\end{equation}

In the final step, we will decide which part will be output based on the current cell state. The output gate determines which part will be output, and then selects the information to be output from the cell state. Figure \ref{fig:output} shows the output gate and the formula is as follows:
\begin{equation}\label{eq:o_t}
	o_t=\sigma(W_o\cdot [h_{t-1},x_t]+b_o)
\end{equation}

\begin{equation}\label{eq:h_t}
	h_t=o_t\cdot tanh(C_t)
\end{equation}

The sequences generated by random walk can be regarded as time series, i.e., there is a temporal correlation among the nodes in the generated sequence. In other words, such sequences reflect the transfer behaviors among the network nodes and can be predicted. Inspired by this, we propose to employ a prediction model, e.g., RNN and LSTM, to process such sequences. The sequence prediction model is a kind of model well-suited to learn from experience to process and predict time series.

The conventional prediction model is popularly applied to natural language processing, mainly for sequence prediction, e.g., predicting the next word or sentence \cite{Sutskever2016} and translating English sentence to another language \cite{Sutskever2014}. In such areas, researchers pay attention to the final result of these tasks. However the network representation learning is concerned with representation of hidden layer \cite{Tian2014}.

For the aim of network representation learning, we suggest a new network structure embedding layer into the conventional prediction network to utilize its sequence prediction ability. The conventional prediction model itself is trained to predict the next step in the sequences \cite{Sun2019Network}. And the embedding layer is used to learn the representation, which maps high-dimensional data to low-dimensional space. Let $g:\mathcal{C}\longmapsto \mathcal{X}$, $\mathcal{X}\in\mathbb{R}^{{\left|V\right|}\times d}, d\ll\left|V\right|$, denotes the embedding layer where $\mathcal{C}$ is the corpus we built by random walk, $\mathcal{X}$ is the $d$-dimensional vector we expected.

For the embedding model training, we use the sequence generated by DW-$RandomWalk$ to predict its next node. It is worth noting that this is a $fake$ task \cite{Mikolov2013}. We just want to obtain the nodes' representation through the predictive procedure. The embedding layer is the real output we expect.

For network data, our goal is to maximize the likelihood of predicting the next node, i.e,:
\begin{equation}\label{equ:p1}
  \mathcal{P}(X^{t+1}|X^t)
\end{equation}
where $X^t = (x^1, x^2, \dots, x^t), X^{t+1} = (x^2, x^3, \dots, x^{t+1})$. We add a $softmax$ function as $softmax(x_i) = \frac{\mathit{e}^{x_i}}{\begin{matrix} \sum_{i} \mathit{e}^{x_i}\end{matrix}}$ after the output layer. For each input sequence, the goal is to predict its next node, so we use cross-entropy to evaluate the loss:
\begin{equation}\label{equ:loss}
\mathcal{L} = -\sum \mathcal{Y}\ln \mathcal{O}
\end{equation}
where $\mathcal{Y}$ is the predicted target node, and $\mathcal{O}$ is the output of the $softmax$ layer.

We use Adaptive Moment Estimation (Adam) \cite{Kingma2014} to optimize the objective function and Back-Propagation Through Time (BPTT) \cite{Williams1990} to update parameters including the representation $\Phi$.

\subsection{Embedding Space Optimization}
The embedding model is able to capture the global structural information and transition context of the network. Because it not only considers the information at time $t_c$, but also considers the information before $t$, e.g. $t_{c-1}, t_{c-2, \dots}$. However, it does not preserve the local structure of the network. We should guarantee the connected nodes still close with each other in the new embedding space. To this end, we propose a Laplacian supervised Embedding space Optimization (LapEO) to preserve the local network structure. The idea is motivated by Laplacian Eigenmaps\cite{Belkin2001}, which attempts to make connected nodes as close as possible in low-dimensional space. Network representation learning is to learn a function to map the data in high-dimensional space to low-dimensional space and maintains structural consistency \cite{Grover2016}. In other words, the neighborhood context of each node in the original space and the new space should be as similar as possible. So we propose the loss function of the Laplacian optimization as follow.

\begin{equation}\label{equ:laploss}
\mathcal{L}_{reg} = \sum_{i,j} (y_i-y_j)^2A_{ij} = 2\cdot Tr(Y^TLY)
\end{equation}
where $Y \in \mathbb{R}^{{\left|V\right|}\times d}$ is the node representation, $A$ is the adjacency matrix, $L = D - A$ is Laplacian matrix, $D \in \mathbb{R}^{n \times n}$ is a diagonal matrix, $D_{i,i} = \sum_j A_{i,j}$.

We optimize the loss functions $\mathcal{L}$ and $\mathcal{L}_{reg}$ in an iterative way. That means we choose an individual way and use different optimization algorithms. There are two main reasons. The first one is that the parameters of RNN and LSTM models are difficult to update when there are two loss functions. The other reason is that the representation is shared between these two stages. The LapEO can be performed after each the embedding model epoch to accelerate the model training procedure.

\begin{algorithm}[h]
\caption{Network Embedding framework via Deep Prediction models (NEDP)}\label{alg}
\begin{algorithmic}[1]
\Require
Network $N(V,E)$, Walks per node $\gamma$, Walk length $l$, Expected dimension $d$
\Ensure
Representation $\Phi \in \mathbb{R}^{{\left|{V}\right|} \times d}$
\State Initialize $walks$ to empty
\State Calculate the normalized transition probability $p$
\For{each $v_i \in V$, $i = 0$ to $\gamma$}
\State Initialize $walk$ to [$v_i$]
\For{$iter = 1$ to $l$ do}
\State $current = walk[-1]$
\State $Neighbors_{current}$=GetNeighbors(G, $current$)
\State $next$ = AliasSample($Neighbor_{current}$, $p$)
\State Append $next$ to $walk$
\EndFor
\State Append $walk$ to $walks$
\EndFor
\Repeat
\State Initialize representation $\Phi$ randomly;
\State Train Deep Prediction embedding model with $walks$;
\State Update Deep Prediction embedding model including representation $\Phi$;
\State Optimize $\Phi$ via Eq. \ref{equ:laploss};
\Until {converge}

\end{algorithmic}
\end{algorithm}

\subsection{Edge Representations}
The NEDP can learn abundant representations for each node in network using transfer behavior and structure. In fact, we are more interested in edges rather than nodes for some prediction tasks. For example, in link prediction task, we predict whether a link exists between two nodes in a network. Because our DW-random walk is based on the social role between nodes in the network, we use the representations of pairs of nodes to represent the edge by some operators in Table \ref{tab:edge_feat}.

\begin{table*}[!h]\large
	\setlength{\tabcolsep}{3.5mm}
	\centering
	\caption{The operator set for constructing edge feature}\label{tab:edge_feat}
	\begin{tabular}{|c|c|c|}
		\hline
		Operator & Mark & Definition \\
		\hline
		Cascade & $\bigcup$ & $y_{edge}(u,v) = y_{node}(u) \lVert y_{node}(v)$ \\
		\hline
		Average & $\bigoplus$ & $y_{edge}(u,v) = \frac{y_{node}(u) + y_{node}(v)}{2}$ \\
		\hline
		Hadamard & $\bigotimes$ & $y_{edge}(u,v) = y_{node}(u) * y_{node}(v)$ \\
		\hline
		L1 Regularization & $|\cdot|_1$ & $y_{edge}(u,v) = |y_{node}(u) - y_{node}(v)|$ \\
		\hline
		L2 Regularization & $|\cdot|_2$ & $y_{edge}(u,v) = {|y_{node}(u) - y_{node}(v)|}^2$ \\
		\hline
	\end{tabular}
\end{table*}

Given pairs of nodes $u$ and $v$, we define an operator over the corresponding representations $y_{node}(u)$ and $y_{node}(v)$ to generate edge's representation $y_{edge}(u,v)$. We desire our operators to be generally defined for any pairs of nodes, even if there is not edge between two nodes since doing so makes the representations robust for many prediction tasks. As shown in Table \ref{tab:edge_feat}, we provide five operators available for learning edge representations.

\section{Experiments}
In this section, we empirically validate the effectiveness of the proposed algorithm in comparison to various state-of-the-art algorithms. To make the results more convincing, we use ten datasets to conduct five experiments: clustering, visualization, classification, reconstruction and link prediction. The codes and datasets can be found from the website https://github.com/songzenghui/NEDP.
\subsection{Datasets}
\begin{table*}[!h]
\setlength{\tabcolsep}{3.5mm}
\centering
\small
\caption{Network structured data used in our experiments}\label{tab:dataset}
\begin{tabular}{|c|c|c|c|c|c|c|}
\hline
\multicolumn{2}{|c|}{ } & $|V|$  & $|E|$  & average degree & label & task \\
\hline
\multirow{2}{*}{Social network} & BlogCatalog & 10,312 & 333,983 & 64.78 & 39 & classification \\
\cline{2-7}
 & Facebook & 4,039 & 88,234 & 43.69 & No & link prediction \\
 \hline
 Biomedical network & SS-Butterfly & 832 & 86,528 & 208.0 & 10 & reconstruction \\
 \hline
 Collaboration network & ca-HepTh & 9,877 & 25,998 & 12.87 & No & reconstruction \\
 \hline
 \multirow{3}{*}{Citation network} & Cora & 2,708 & 5,278 & 3.90 & 7 & classification \\
 \cline{2-7}
  & Citeseer & 3,327 & 4,676	 & 2.81 & 6 & classification \\
 \cline{2-7}
  & Pubmed & 19,717 & 44,327 & 4.50 & 3 & classification \\
 \hline
 \multirow{3}{*}{20-Newsgroup} & 3-NG & 600 & 179,700 & 599.0 & 3 & cluster\&visualization \\
 \cline{2-7}
  & 6-NG & 1,200 & 719,400 & 1,199.0 & 6 & cluster \\
 \cline{2-7}
  & 9-NG & 1,800 & 1,619,100 & 1,799.0 & 9 & cluster \\ 
 \hline
\end{tabular}
\end{table*}
The following datasets are employed for our experiments. The datasets contain various numbers of nodes and edges which could give sufficient validation for all the methods, especially those large networks, e.g., Pubmed and BlogCatalog.
\begin{itemize}
\item BlogCatalog \cite{Tang2009} is a social network dataset. The node represents the blogger, the edge represents the attention between blogger, and the label represents the interest of the blogger. This dataset has $10,312$ nodes, $333,982$ edges, and $39$ different labels
\item Facebook \cite{Mcauley2012} is a network dataset from social platform. Each node represents a user, while each edge represents a connection between two users. This dataset has $4,039$ nodes, $88,234$ edges. Although the dataset has no labels, it is sampled from the real world and can be used to simulate user-recommended tasks.
\item SS-Butterfly \cite{Bo2018}\cite{Wang2009} is a biomedical network dataset. Nodes represent butterflies and edges represent visual similarities between the organisms. It has $832$ nodes, $86,528$ edges, $10$ labels. 
\item ca-HepTh \cite{Leskovec2007} is a collaboration network from the arXiv website. The nodes represent authors and the edges represent the relationship between two authors who have completed a paper together. This dataset has $9,877$ nodes and $25,998$ edges.
\item Cora \cite{Sen2008} is a citation network dataset. The nodes represent papers, the edges represent citations, the labels represent conferences published by papers.This dataset has 2,708 nodes, 5,429 edges, and 7 different labels.
\item Citeseer \cite{Sen2008} is a citation network. The nodes represent papers, the edges represent citations, the labels represent the field to which the paper belongs. This dataset has 3,327 nodes, 4,732 edges, and 6 different labels.
\item Pubmed \cite{Sen2008} is a Biomedical papers citation network dataset. The nodes represent papers, edges represent citations, labels represent conferences. This dataset has 19,717 nodes, 44,338 edges, 3 different labels.
\item The 20-Newsgroups \cite{Lang1995} dataset is a collection of 20,000 newsgroup documents, partitioned into 20 different categories. In this dataset, we use a vector with TF-IDF scores of each word to represent the documents, and calculate the cosine similarity to represent the similarity between two documents. Based on these similarity scores of each pair of documents, we build a full-connection language network. According to \cite{Tian2014}, we randomly sample 200 documents from each category to make three networks built from 3, 6 and 9 different newsgroups respectively and name them 3-NG, 6-NG and 9-NG (NG refers to the newsgroup).
\end{itemize}

\subsection{Baseline Methods}
We compare our method with several baseline methods, including DeepWalk \cite{Perozzi2014}, LINE \cite{Tang2015}, SDNE \cite{Wang2016}, Struc2Vec \cite{Ribeiro2017}, and GraphGAN  \cite{Wang2017}. There are also some other embedding methods, however, we can not show all of them here. The above methods are recently proposed and give extensive experiments with other methods in the corresponding papers.

\subsection{Parameter Settings}
For the compared methods, we set the optimal parameters as suggested in their original papers. For example, DeepWalk, we set window size as $10$, walk length as $40$ and walks per vertex as $40$. For LINE, the number of negative samples is set as $5$ and the total number of samples is $10$ billion. For SDNE, we set the number of layers in the model as $3$, the hyper-parameter $\alpha$ and $\beta$ as $0.1$ and $10$. For Struc2Vec, we set window size as $10$, walk length as $80$, walks per vertex as $10$. All methods get representation with dimensions of $128$. For our method, we set walk length to be $100$. We use different $\gamma$ ( walks per node) for different datasets. For BlogCatalog and Pubmed datasets, we set walks per node as 30. For other datasets, we set walks per node as $100$. The LSTM learning rate is 0.001. For convenience, we set LSTM timesteps equal to walk length $l$.

\subsection{Experiments Results}
In this section, we evaluate the learned representations of different methods through five downstream tasks: clustering, visualization, classification, reconstruction and link prediction.
\subsubsection{Clustering}
\begin{table}[h]\large
	\setlength{\tabcolsep}{4mm}
	\centering
	\small
	\caption{The result of clustering}\label{tab:cluster}
	\begin{tabular}{|c|c|c|c|}
		\hline
		& 3-NG & 6-NG & 9-NG \\
		\hline
		DeepWalk & 0.351 & 0.300 & 0.197 \\
		\hline
		LINE & 0.625 & 0.461 & 0.215 \\
		\hline
		SDNE & 0.423 & 0.365 & 0.254 \\
		\hline
		GraphGAN & 0.543 & 0.315 & 0.201 \\
		\hline
		Struc2Vec & 0.412 & 0.307 & 0.211 \\
		\hline
		NEDP-RNN & 0.639 & 0.520 & 0.321 \\
		\hline
		NEDP-LSTM & \textbf{0.709} & \textbf{0.527} & \textbf{0.425} \\
		\hline
	\end{tabular}
\end{table}
Most of the data in the real world are unlabeled, so learning the representation is particularly important for unsupervised learning task. In order to evaluate the effectiveness of our representation in unsupervised learning tasks, we design a clustering experiment on three datasets: 3-NG, 6-NG and 9-NG, which are separated from 20-Newsgroup.

In the clustering task, we execute each baseline method to generate representations for the nodes, which are used as features for clustering. Then we cluster the nodes into categories with K-Means algorithm. And we record the performance with NMI (Normalized Mutual Information) score. The results of clustering are shown in Table \ref{tab:cluster}

\begin{figure*}[t]
	\centering
	\subfloat[NEDP-LSTM]{\includegraphics[width=0.14\textwidth]{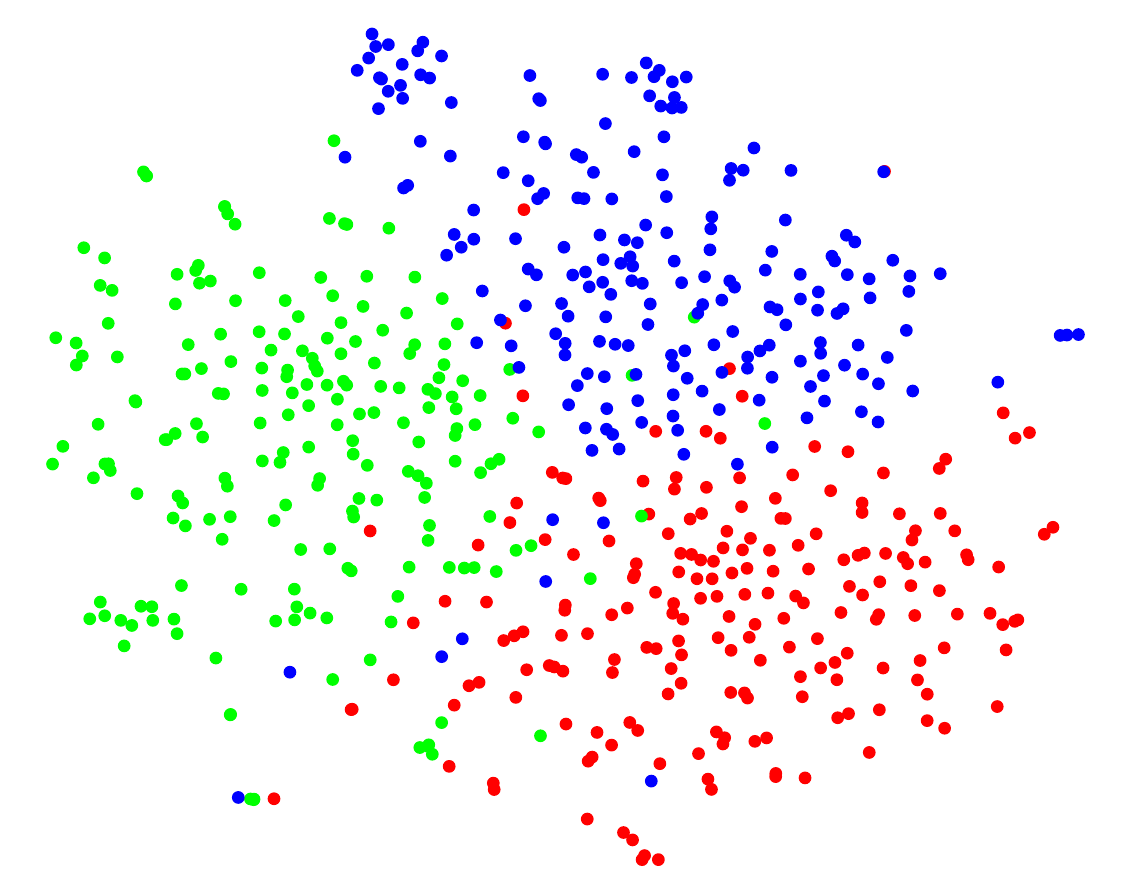}\label{fig:vis_our}}
	\subfloat[NEDP-RNN]{\includegraphics[width=0.14\textwidth]{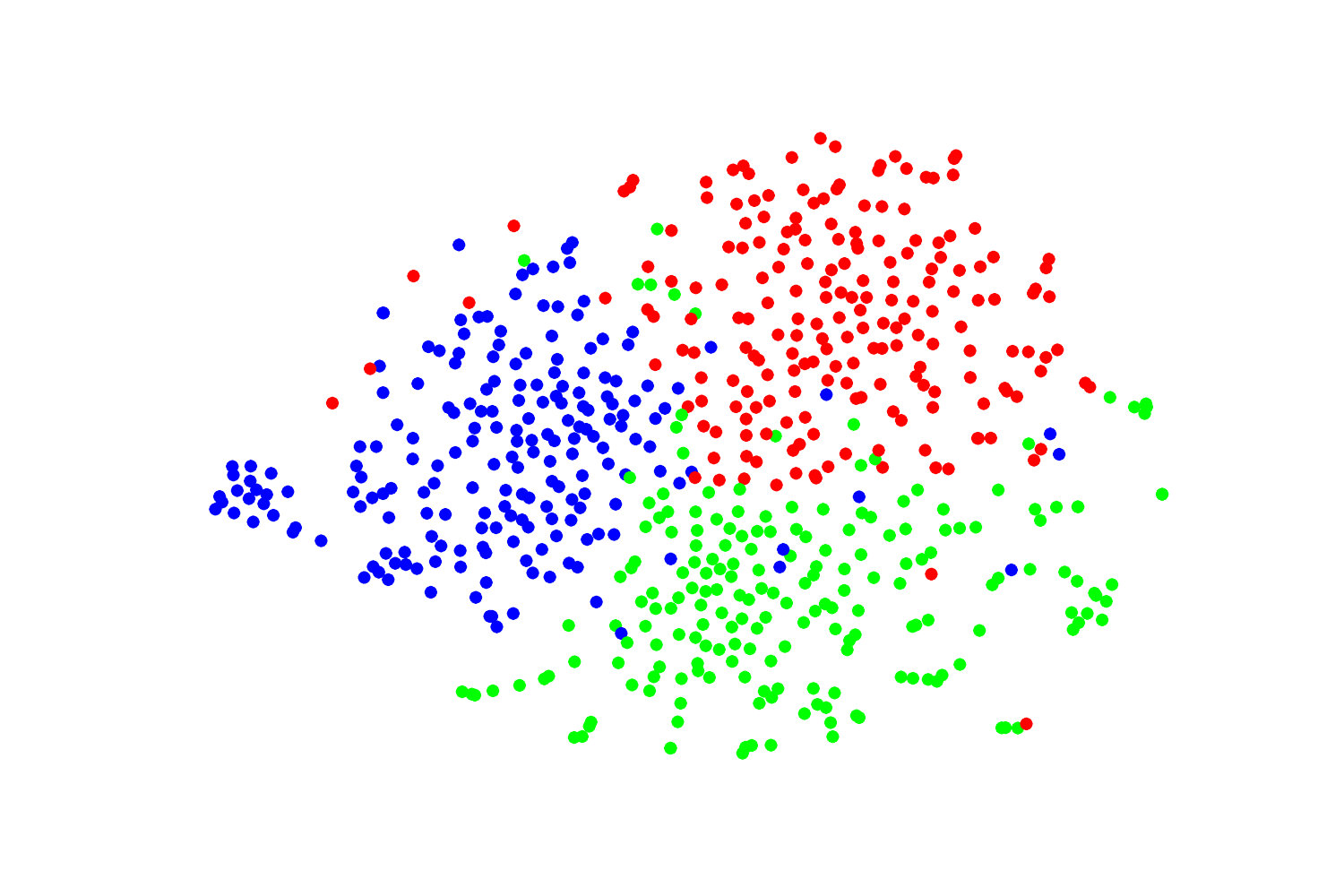}\label{fig:vis_rnn}}
	\subfloat[DeepWalk]{\includegraphics[width=0.14\textwidth]{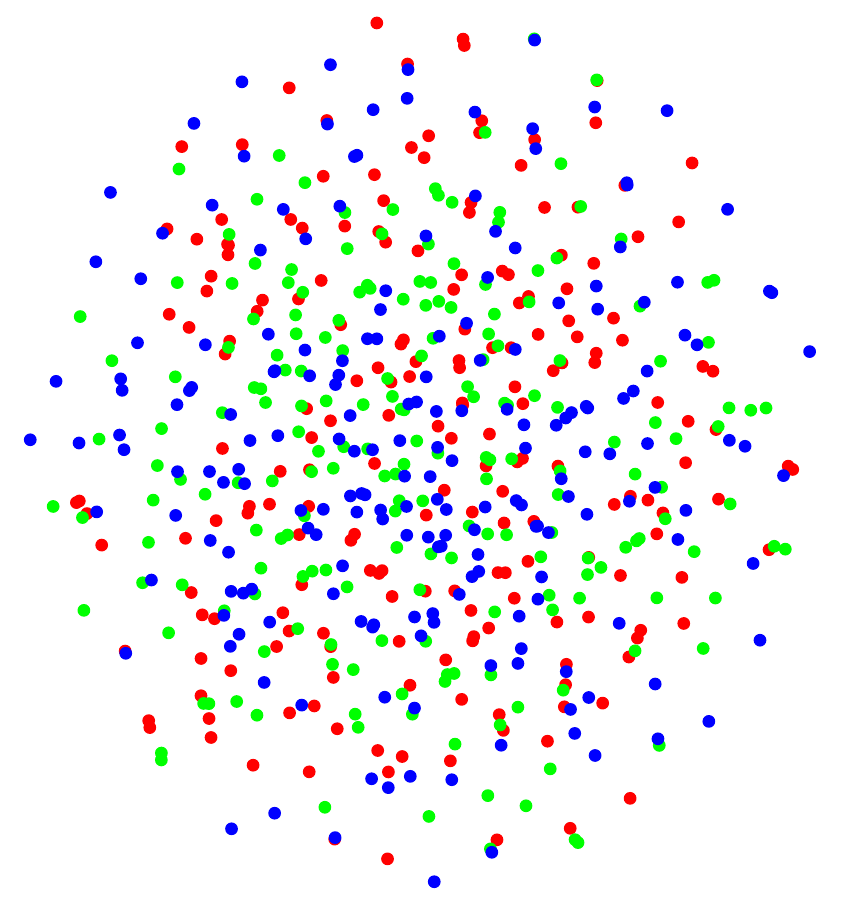}\label{fig:vis_dw}}
	\subfloat[LINE]{\includegraphics[width=0.14\textwidth]{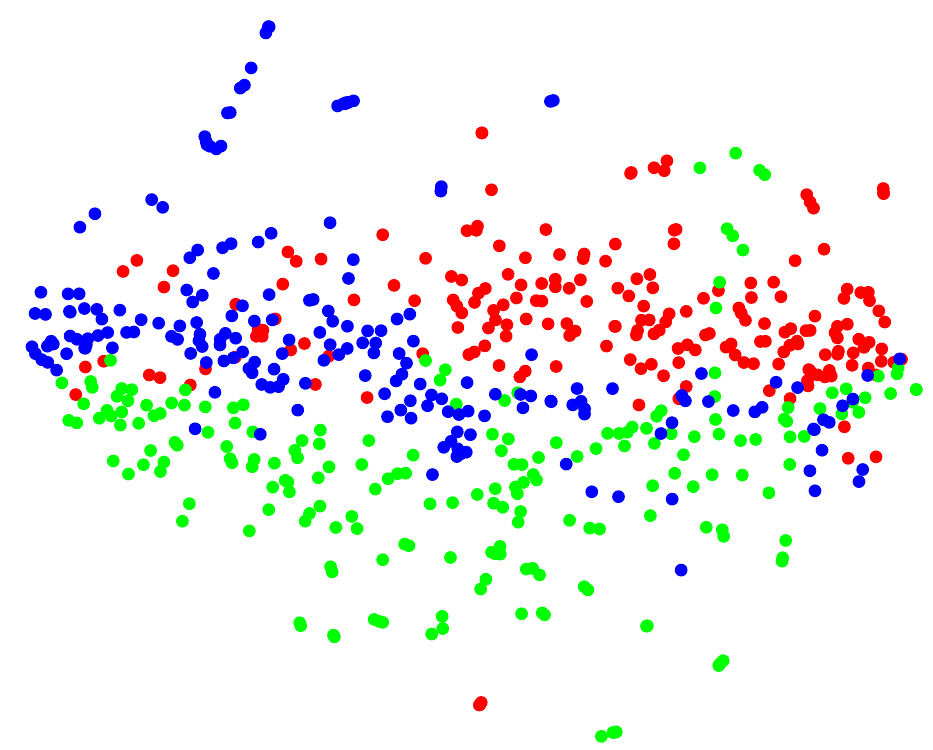}\label{fig:vis_line}}
	\subfloat[SDNE]{\includegraphics[width=0.14\textwidth]{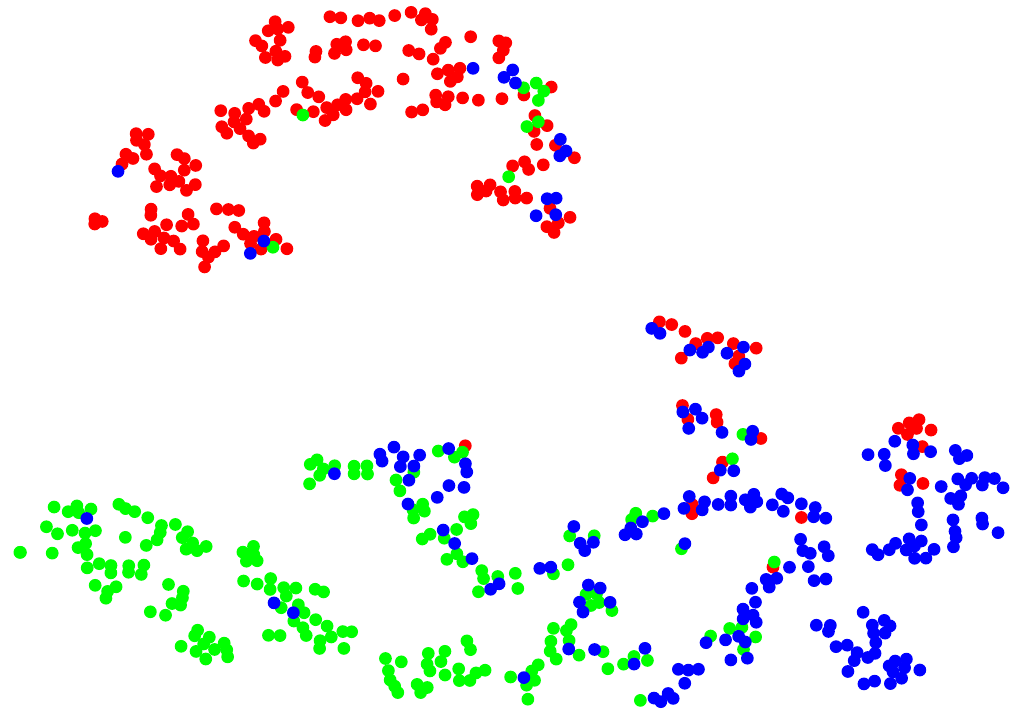}\label{fig:vis_sdne}}
	\subfloat[GraphGAN]{\includegraphics[width=0.14\textwidth]{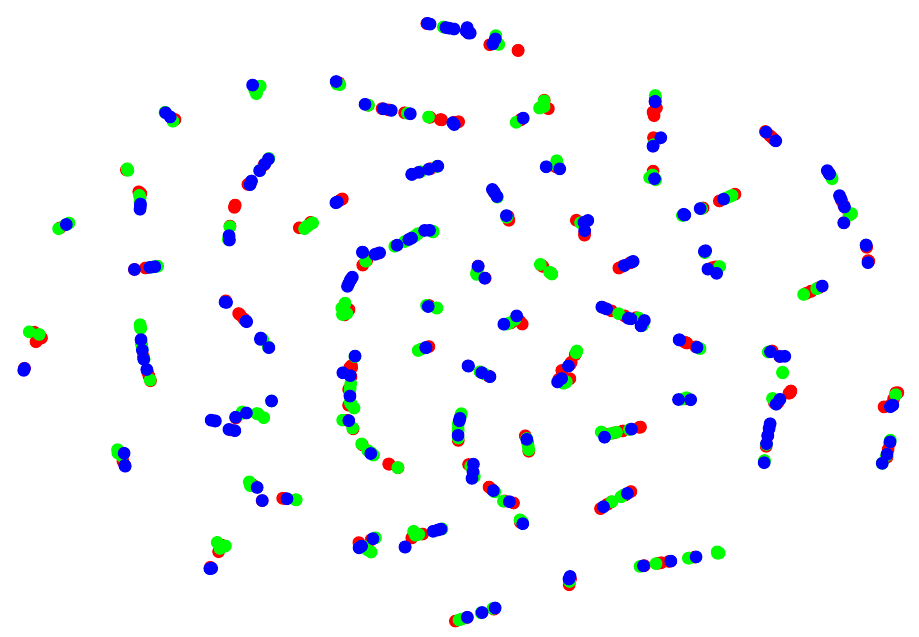}\label{fig:vis_gg}}
	\subfloat[Struc2Vec]{\includegraphics[width=0.14\textwidth]{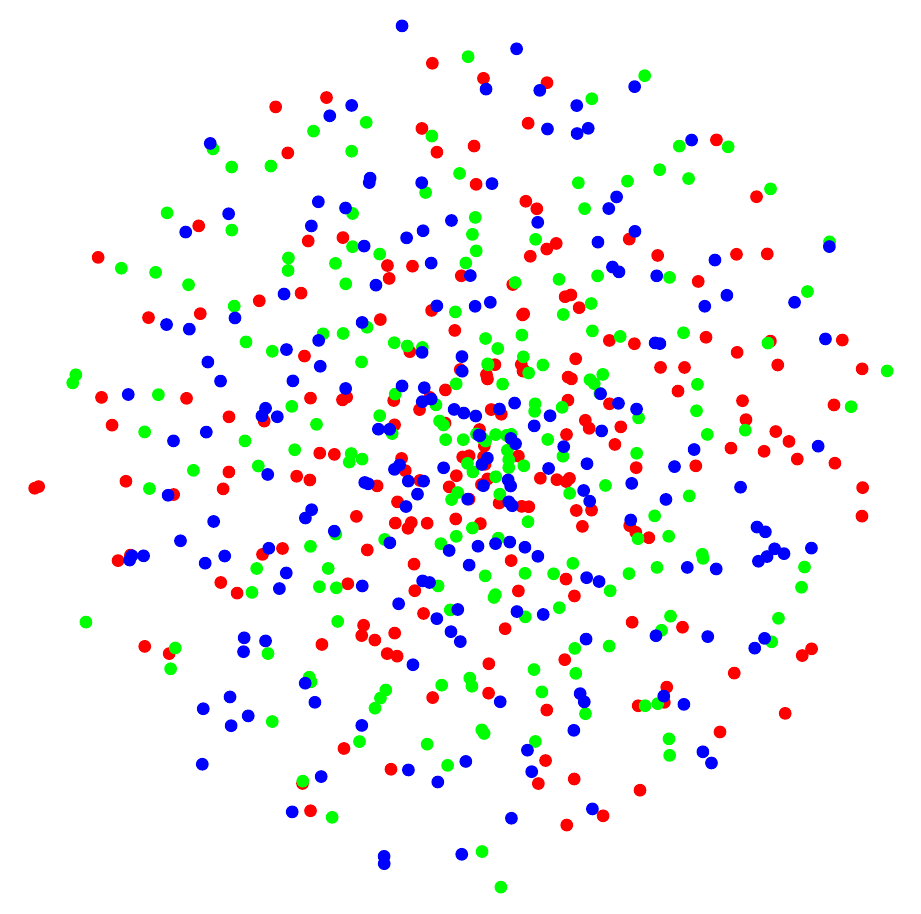}\label{fig:vis_s2v}}
	\caption{Visualization of 3-NG dataset. Each point represents one document. Different colors correspond to different categories, i.e., Red: $comp.graphics$, Blue: $rec.sport.baseball$, Green: $talk.politics.guns$ }
	\label{fig:vis}
\end{figure*}

The results show that our method outperforms the others. Methods, such as DeepWalk and GraphGAN, only consider whether two nodes are connected and do not take the weight of edges into account. Therefore, these baselines are not applicable to the weighted dense networks. However, our method overcomes these obstacles. The proposed DW-random walk method not only considers the connection between two nodes, but also the weights of edges and the degrees of nodes. Note that NEDP-RNN's performance is second only to NEDP-LSTM's, which illustrates that LSTM model precedes RNN model for its long term dependency. In combination with LSTM and LapEO, we can better preserve the network's global and local information. Therefore NEDP-LSTM is robust in the clustering task both on a weighted and unweighted dense network.

\subsubsection{Visualization}

In visualization task, we focus on using the learned representation to reveal the network data intuitively. We execute our model and baseline methods on 3-NG dataset which comes from the 20-Newsgroup dataset. This dataset has 600 nodes,  each of which belongs to one of three categories which are $comp.graphics$, $rec.sport.baseball$ and $talk.politics.guns$. We map the representations learned by different network embedding methods into the 2-D space using the visualization tool $t$-$SNE$\cite{Van2017}. Figure \ref{fig:vis} shows the visualization results on 3-NG dataset. Each point represents a document and colors indicate different categories. The visualizations of DeepWalk, Struc2Vec and GraphGAN is not meaningful, where the documents belonging to the same categories are not clustered together. For example, DeepWalk and Struc2Vec make the points belonging to different categories mix with each other. GraphGAN overlaps the nodes of different categories with each other. For LINE and SDNE, although the data can generally be divided into three clusters, the boundary is not clear enough. Obviously, the visualization of our method NEDP including NEDP-RNN and NEDP-LSTM performs better than baselines. This experiment demonstrates the NEDP model can learn more meaningful and robust representations.

\subsubsection{Classification}
\begin{table*}[!h]\normalsize

	\centering
	\caption{The result of Multilabel classification on BlogCatalog}\label{tab:blog}
	\begin{tabular}{|c|c|c|c|c|c|c|c|c|c|c|}
		\hline
		& \% Labeled Nodes & 10\% & 20\% & 30\% & 40\% & 50\% & 60\% & 70\% & 80\% & 90\% \\
		\hline
		
		\multirow{5}*{Micro-F1(\%)} & DeepWalk & 33.12 & 36.20 & 37.60 & 39.30 & \textbf{40.00} & 40.30 & 40.50 & \textbf{41.50} & 42.00 \\
		& LINE & 31.03 & 33.41 & 34.56 & 35.42 & 35.97 & 36.18 & 36.74 & 36.82 & 36.89 \\
		& SDNE & 30.81 & 31.94 & 32.56 & 32.98 & 33.12 & 33.31 & 33.47 & 33.57 & 33.95 \\
		& GraphGAN & 23.43 & 24.83 & 25.30 & 25.62 & 25.67 & 25.51 & 25.69 & 25.73 & 25.34 \\
		& Struc2Vec & 10.73 & 11.63 & 12.57 & 13.24 & 13.86 & 14.25 & 14.56 & 14.87 & 14.56 \\
		& NEDP-RNN & 31.52 & 34.47 & 36.63 & 37.93 & 38.72 & 39.16 & 39.70 & 39.97 & 40.47 \\
		& NEDP-LSTM & \textbf{34.30} & \textbf{37.13} & \textbf{37.91} & \textbf{39.39} & 39.97 & \textbf{40.66} & \textbf{40.88} & 41.32 & \textbf{42.13} \\
		\hline
		\multirow{5}*{Macro-F1(\%)}& DeepWalk &\textbf{ 17.79} & 20.02 & 21.62 & 22.57 & 23.18 & 24.66 & 24.73 & 25.11 & 27.37 \\
		& LINE & 11.97 & 14.62 & 16.00 & 17.22 & 18.20 & 18.92 & 19.43 & 19.95 & 20.43 \\
		& SDNE & 14.94 & 15.55 & 16.03 & 16.11 & 16.31 & 16.41 & 16.54 & 16.69 & 16.87 \\
		& GraphGAN & 9.53 & 10.08 & 10.32 & 10.44 & 10.38 & 10.27 & 10.44 & 10.44 & 9.88 \\
		& Struc2Vec & 5.13 & 5.19 & 5.09 & 5.07 & 5.00 & 4.67 & 4.53 & 4.31 & 4.11 \\
		&NEDP-RNN & 17.44 & 20.44 & 22.34 & 23.50 & 24.14 & 24.73 & 25.10 & 25.34 & 25.94 \\
		& NEDP-LSTM & 17.52 & \textbf{21.54} & \textbf{22.88} & \textbf{23.58} & \textbf{24.32} & \textbf{24.90} & \textbf{25.49} & \textbf{25.64} & \textbf{28.37} \\
		\hline
	\end{tabular}
\end{table*}

\begin{table*}[!h]\normalsize
	\centering
	\caption{The result of Node Classification on Cora}\label{tab:cora}
	\begin{tabular}{|c|c|c|c|c|c|c|c|c|c|c|}
		\hline
		& \% Labeled Nodes & 10\% & 20\% & 30\% & 40\% & 50\% & 60\% & 70\% & 80\% & 90\% \\
		\hline
		\multirow{5}*{Accuracy(\%)}& DeepWalk & 71.77 & 73.85 & 74.98 &\textbf{76.12} & \textbf{77.65} & 77.54 & \textbf{78.84} & 79.17 & 80.40 \\
		& LINE & 46.02 & 51.18 & 54.00 & 55.22 & 56.38 & 56.89 & 57.24 & 57.97 & 59.29 \\
		& SDNE & 38.89 & 39.30 & 39.26 & 38.85 & 38.72 & 38.64 & 38.92 & 38.70 & 38.67 \\
		& GraphGAN & 24.58 & 26.64 & 27.44 & 27.75 & 28.72 & 29.33 & 29.42 & 29.62 & 28.71 \\
		& Struc2Vec & 31.67 & 34.19 & 35.87 & 36.91 & 38.20 & 38.53 & 39.59 & 40.40 & 40.33 \\
		& NEDP-RNN & 67.88 & 71.44 & 72.90 & 73.85 & 74.35 & 74.77 & 74.94 & 75.53 & 77.01 \\
		& NEDP-LSTM & \textbf{71.86} & \textbf{74.38} & \textbf{75.12} & 76.06 & 77.03 & \textbf{78.87} & 78.22 & \textbf{79.89} & \textbf{81.18} \\
		\hline
	\end{tabular}	
\end{table*}

\begin{table*}[!h]\normalsize
	\centering
	\caption{The result of Node Classification on Citeseer}\label{tab:citeseer}
	\begin{tabular}{|c|c|c|c|c|c|c|c|c|c|c|}
		\hline
		& \% Labeled Nodes & 10\% & 20\% & 30\% & 40\% & 50\% & 60\% & 70\% & 80\% & 90\% \\
		\hline
		\multirow{5}*{Accuracy(\%)} & DeepWalk & 50.16 & 52.42 & 54.53 & 55.53 & 56.12 & 57.02 & 57.68 & 58.63 & 59.42 \\
		& LINE & 32.75 & 35.29 & 36.84 & 37.72 & 38,12 & 38.65 & 39.30 & 38.98 & 41.10 \\
		& SDNE & 31.27 & 32.90 & 33.56 & 33.53 & 33.63 & 33.94 & 34.34 & 34.09 & 34.77 \\
		& GraphGAN & 19.89 & 20.74 & 20.96 & 21.16 & 21.24 & 21.60 & 21.53 & 21.39 & 22.46 \\
		& Struc2Vec & 25.92 & 26.94 & 27.99 & 28.56 & 28.50 & 29.64 & 30.26 & 30.33 & 31.92 \\
		& NEDP-RNN & 45.32 & 47.56 & 49.63 & 51.27 & 52.32 & 53.24 & 53.70 & 54.44 & 55.46 \\
		& NEDP-LSTM & \textbf{51.72} & \textbf{54.32} & \textbf{55.47} & \textbf{56.08} & \textbf{57.15} & \textbf{58.22} & \textbf{59.75} & \textbf{59.45} & \textbf{61.56} \\
		\hline
	\end{tabular}	
\end{table*}

\begin{table*}[!h]\normalsize
	\centering
	\caption{The result of Node Classification on Pubmed}\label{tab:pubmed}
	\begin{tabular}{|c|c|c|c|c|c|c|c|c|c|c|}
		\hline
		& \% Labeled Nodes & 10\% & 20\% & 30\% & 40\% & 50\% & 60\% & 70\% & 80\% & 90\% \\
		\hline
		\multirow{5}*{Accuracy(\%)} & DeepWalk & 70.75 & 72.29 & 73.47 & 73.59 & 74.58 & 75.71 & 75.69 & 75.83 & 76.65 \\
		&LINE & 53.49 & 54.43 & 54.96 & 55.19 & 55.43 & 55.56 & 55.68 & 55.39 & 55.18 \\
		& SDNE & 39.57 & 40.02 & 40.52 & 40.45 & 41.20 & 41.58 & 41.64 & 41.56 & 41.69 \\
		& GraphGAN & - & - & - & - & - & - & - & - & - \\
		& Struc2Vec & 47.84 & 49.12 & 49.61 & 49.85 & 49.93 & 50.06 & 50.16 & 50.38 & 50.73 \\
		& NEDP-RNN & 70.09 & 71.45 & 71.89 & 72.09 & 72.13 & 72.23 & 72.30 & 72.45 & 73.36 \\
		& NEDP-LSTM & \textbf{73.74} & \textbf{74.45} & \textbf{75.40} & \textbf{75.42} & \textbf{75.57} & \textbf{76.09} & \textbf{76.43} & \textbf{76.70} & \textbf{77.23} \\
		\hline
	\end{tabular}	
\end{table*}

We divide the classification into two categories according to the number of labels in datasets that nodes have: multi-label classification which the node has more than one label and multiclass classification which the node only has one definite label.In the multi-label classification task, i.e., BlogCatalog, every blogger author is assigned one or more interesting topics as labels. The training phase learns a classifier with a certain fraction of nodes and all their labels. The test task is to predict the labels for the remaining nodes. In multi-label classification experiments, we randomly sample a portion (from 10\% to 90\%) of the labeled nodes, and use them as training data. The rest of the nodes are used as test data. We report the average performance in terms of $Micro$-$F_1$ and $Macro$-$F_1$. The results are shown in Tabel \ref{tab:blog}. As we can see that although the performance of DeepWalk is very competitive, we still outperform it. It demonstrates that LSTM deep learning model more powerful than Skip-Gram model. Our model achieves most gains of 27.57\%($Micro$-$F_1$) and 24.26\%($Macro$-$F_1$) with 90\% train nodes. That illustrates that the learned network representations of NEDP can better generalize to the multilabel classification task than baselines.

In single-label node classification task, we employ Cora, Citeseer and Pubmed datasets. And we record the performance in terms of $Accuracy$. For classification experiment, we execute our model and baseline methods on the whole network to generate nodes representations for a one-vs-rest logistic regression classifier. We randomly sample 10\% to 90\% of the nodes as the training samples and use the left nodes to test the performance. The results are shown in Tabel \ref{tab:cora}, \ref{tab:citeseer}, and \ref{tab:pubmed} respectively. For the PubMed dataset, we cannot get a result for the GraphGAN method with the current computational resources. From the results, it is evident we can see NEDP achieves promising performance compared with others. For example, on CiteSeer, our method gives about 2\% gain in accuracy over DeepWalk (ranked in the second place) under all training ratio settings.

\subsubsection{Network Reconstruction}
\begin{figure}[ht]
	\centering
	\includegraphics[scale=0.5]{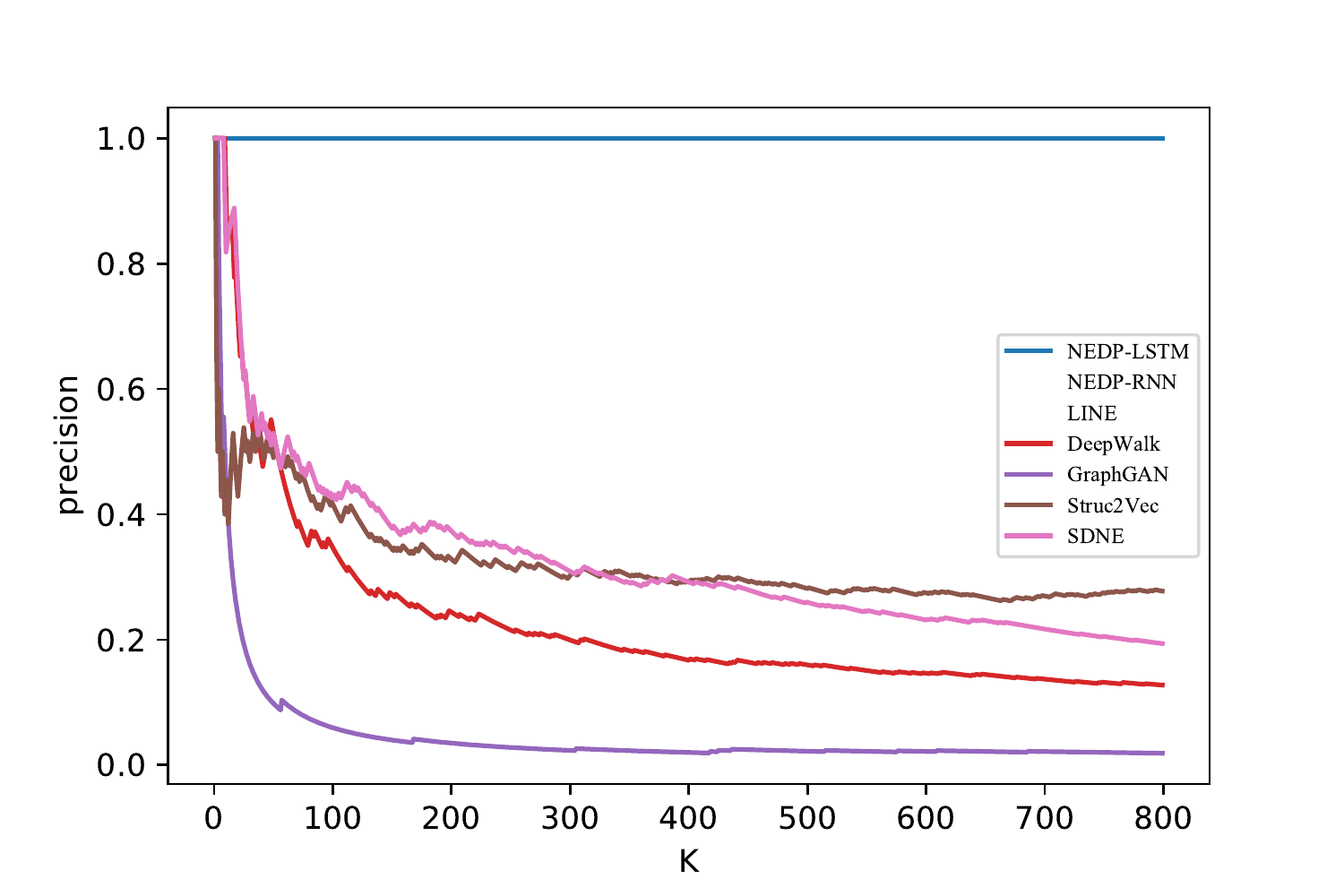}
	\caption{The result on SS-Butterfly dataset. Three lines overlap because the results of three methods are the same.}
	\label{fig:bf}
\end{figure}

\begin{figure}[ht]
	\centering
	\includegraphics[scale=0.5]{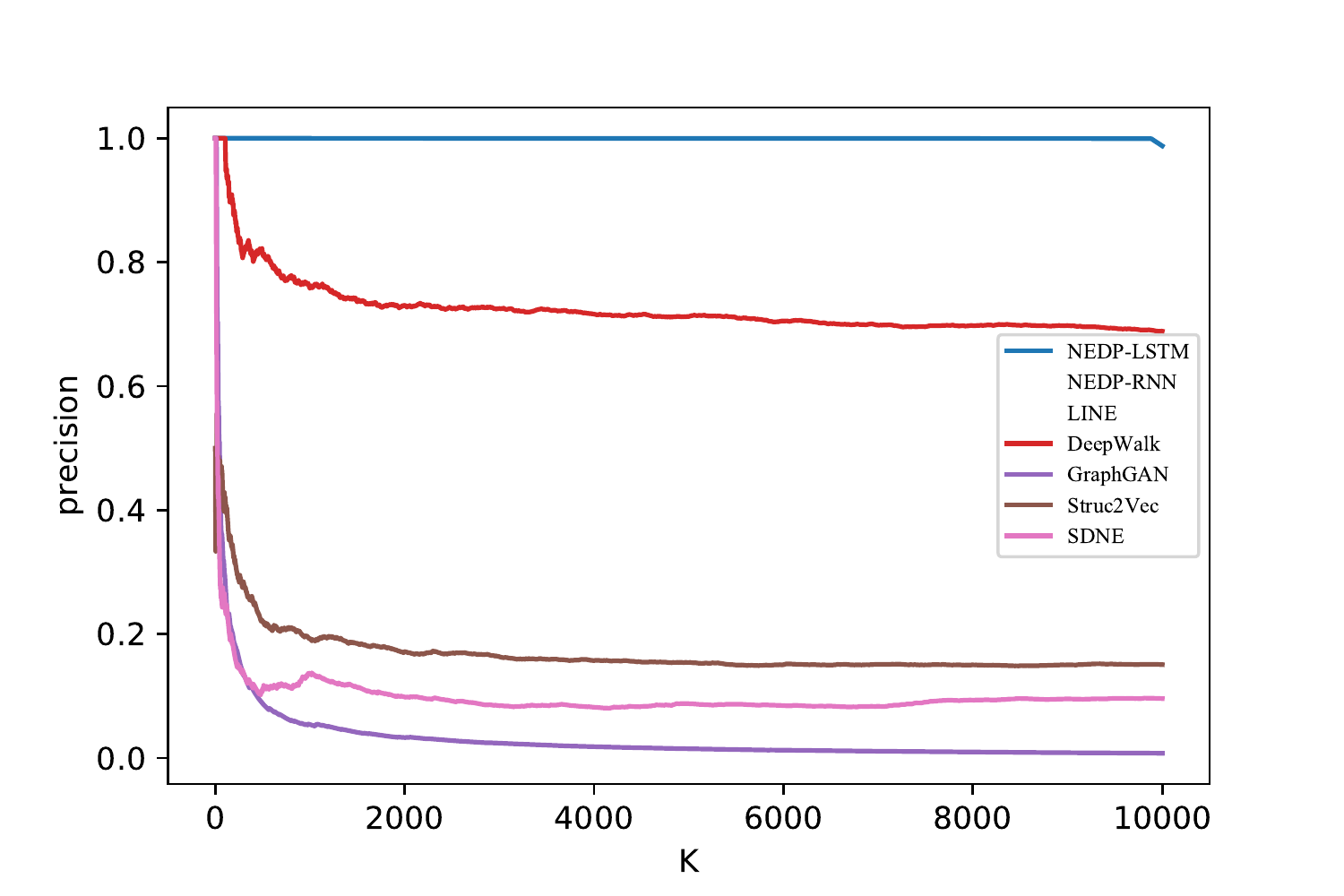}
	\caption{The result on ca-HepTh dataset. Two lines overlap because the results of two methods are the same.}
	\label{fig:hepth}
\end{figure}

To verify that the representation learned from embedding methods can preserve the network structure, we conduct the network reconstruction experiment that is a basic evaluation on network embedding learning methods according to their performance. The experiments are carried out on a biomedical network SS-Butterfly and a collaboration network ca-HepTh, which both have a large number of edges. The node representations learned by network representation learning methods keep the edge features for network reconstruction. Excellent representations can restore the original network structure by measuring similarity in their vector space. Given a network, we use different network representation learning methods to learn the network representations and rank pairs of nodes according to their similarities. The highest $K$ ranking pairs of nodes are used to reconstruct the network. We use the $precision@K$ and $MAP$ as the evaluation metrics. Among them, $precision@K$ can reflect the single local prediction result while $MAP$ can reflect the multiple global prediction result. The results on SS-Butterfly and ca-HepTh dataset are presented in Figure \ref{fig:bf} and Figure \ref{fig:hepth} respectively. The result of $MAP$ is shown in Table \ref{tab:recon}.

\begin{table*}[!h]\normalsize
	\centering
	\caption{The result of Network Reconstruction on SS-Butterfly and ca-HepTh dataset}\label{tab:recon}
	\begin{tabular}{|c|c|c|c|c|c|c|c|}
		\hline
		 & DeepWalk & LINE & SDNE & GraphGAN & Struc2Vec & NEDP-RNN & NEDP-LSTM \\
		 \hline
		 SS-Butterfly & 0.2271 & \textbf{1.0} & 0.3238 & 0.0468 & 0.3229 & \textbf{1.0} & \textbf{1.0} \\
		 \hline
		 ca-HepTh & 0.7235 & \textbf{1.0} & 0.0983 & 0.0307 & 0.1684 & \textbf{1.0} & 0.9999 \\
		 \hline
	\end{tabular}
\end{table*}

From Fig. \ref{fig:bf} and Fig. \ref{fig:hepth} we can observe that NEDP-LSTM, NEDP-RNN and LINE all get good results which achieve around $100\%$ and remain $100\%$ with the increase of $K$. This indicates that our method can effectively reconstruct the original network, even the network has fewer nodes and more edges. LINE model can reconstruct the original network perfectly, which benefits from both the first-order proximity and the second-order proximity. It shows that preserving the structural information of the network is important. Table \ref{tab:recon} shows that our method achieves competitive performance in $MAP$ with LINE model in both datasets. It demonstrates that our method is able to preserve the network structure well.

\subsubsection{Link Prediction}
\begin{table*}[!h]\normalsize
	\centering
	\caption{The result of Link Prediction on Facebook (removed 15\% edges)}\label{tab:lp15}
	\begin{tabular}{|c|c|c|c|c|c|c|c|c|}
		\hline
		& & DeepWalk & LINE & SDNE & GraphGAN & Struc2Vec & NEDP-RNN & NEDP-LSTM \\
		\hline
		$\bigcup$ & AUC & 0.7868 & 0.7725 & 0.8017 & 0.7988 & 0.8124 & \textbf{0.8225} & \textbf{0.8133} \\
		\cline{2-9}
		& AP & 0.8099 & 0.7908 & 0.8077 & 0.8141 & 0.8172 & \textbf{0.8330} & \textbf{0.8232} \\
		\hline
		$\bigoplus$ & AUC & 0.7859 & 0.7704 & 0.8012 & 0.7976 & 0.8119 & \textbf{0.8222} & \textbf{0.8129} \\
		\cline{2-9}
		& AP & 0.8136 & 0.7905 & 0.8082 & 0.8140 & 0.8166 & \textbf{0.8333} & \textbf{0.8225} \\
		\hline
		$\bigotimes$ & AUC & \textbf{0.9842} & 0.9439 & \textbf{0.9843} & 0.9222 & 0.7039 & 0.9496 & 0.9405 \\
		\cline{2-9}
		& AP & \textbf{0.9779} & 0.9473 & \textbf{0.9843} & 0.9429 & 0.7195 & 0.9590 & 0.9278 \\
		\hline
		$|\cdot|_1$ & AUC & \textbf{0.9881} & 0.9309 & 0.9766 & 0.8154 & 0.6809 & \textbf{0.9827} & 0.9498 \\
		\cline{2-9}
		& AP & \textbf{0.9892} & 0.9346 & 0.9784 & 0.8546 & 0.6780 & \textbf{0.9847} & 0.9546 \\
		\hline
		$|\cdot|_2$ & AUC & \textbf{0.9886} & 0.9377 & 0.9762 & 0.8163 & 0.6837 & \textbf{0.9837} & 0.9532 \\
		\cline{2-9}
		& AP & \textbf{0.9894} & 0.9404 & 0.9775 & 0.8548 & 0.6798 & \textbf{0.9855} & 0.9577 \\
		\hline
	\end{tabular}
\end{table*}

\begin{table*}[!h]\normalsize
	\centering
	\caption{The result of Link Prediction on Facebook (removed 30\% edges)}\label{tab:lp30}
	\begin{tabular}{|c|c|c|c|c|c|c|c|c|}
		\hline
		& & DeepWalk & LINE & SDNE & GraphGAN & Struc2Vec & NEDP-RNN & NEDP-LSTM \\
		\hline
		$\bigcup$ & AUC & 0.7813 & 0.7759 & 0.8092 & 0.7981 & \textbf{0.8092} & \textbf{0.8196} & 0.8001 \\
		\cline{2-9}
		& AP & 0.8045 & 0.7925 & \textbf{0.8139} & 0.8100 & 0.8124 & \textbf{0.8325} & 0.8064 \\
		\hline
		$\bigoplus$ & AUC & 0.7802 & 0.7751 & 0.8083 & 0.7972 & \textbf{0.8086} & \textbf{0.8194} &	0.7995 \\
		\cline{2-9}
		& AP & 0.8074	 & 0.7929 & 0.8144 & \textbf{0.8144} & 0.8117 & \textbf{0.8331} & 0.8063 \\
		\hline
		$\bigotimes$ & AUC & \textbf{0.9837} & 0.9321 & \textbf{0.9819} & 0.8697 & 0.7001 & 0.9441	& 0.9524 \\
		\cline{2-9}
		& AP & \textbf{0.9763} & 0.9379 & \textbf{0.9819} & 0.9007 & 0.7141 & 0.9482 & 0.9563 \\
		\hline
		$|\cdot|_1$ & AUC & \textbf{0.9871} & 0.9239 & 0.9729 & 0.7324 & 0.6775 & \textbf{0.9805}	& 0.9521 \\
		\cline{2-9}
		& AP & \textbf{0.9878} & 0.9239 & 0.9737 & 0.7877 & 0.6787 & \textbf{0.9826} & 0.9550 \\
		\hline
		$|\cdot|_2$ & AUC & \textbf{0.9873} & 0.9296 & 0.9712 & 0.7332 & 0.6880 & \textbf{0.9823} & 0.9575 \\
		\cline{2-9}
		& AP & \textbf{0.9879} & 0.9243 & 0.9712 & 0.7794 & 0.6910 & \textbf{0.9814} & 0.9601 \\
		\hline
	\end{tabular}
\end{table*}

\begin{table*}[!h]\normalsize
	\centering
	\caption{The result of Link Prediction on Facebook (removed 50\% edges)}\label{tab:lp50}
	\begin{tabular}{|c|c|c|c|c|c|c|c|c|}
		\hline
		& & DeepWalk & LINE & SDNE & GraphGAN & Struc2Vec & NEDP-RNN & NEDP-LSTM \\
		\hline
		$\bigcup$ & AUC & 0.7792 & 0.7729 & 0.8068 & 0.7769 & \textbf{0.8255} & \textbf{0.8171} & 0.8050 \\
		\cline{2-9}
		& AP & 0.8040 & 0.7902 & 0.8113 & 0.8012 & \textbf{0.8328} & \textbf{0.8303} & 0.8039 \\
		\hline
		$\bigoplus$ & AUC & 0.7782 & 0.7722 & 0.8059 & 0.7766 & \textbf{0.8253} & \textbf{0.8167} & 0.8042 \\
		\cline{2-9}
		& AP & 0.8068 & 	0.7901 & 0.8108 & 0.8018 & \textbf{0.8228} & \textbf{0.8312} & 0.8054 \\
		\hline
		$\bigotimes$ & AUC & \textbf{0.9797} & 0.9125 & \textbf{0.9736} & 0.7908 & 0.7409 & 0.9184 & 0.9192 \\
		\cline{2-9}
		& AP & \textbf{0.9710} & 0.9093 & \textbf{0.9758} & 0.8363 & 0.7510 & 0.9287 & 0.9137 \\
		\hline
		$|\cdot|_1$ & AUC & \textbf{0.9831} & 0.8963 & 0.9618 & 0.6554 & 0.7204 & \textbf{0.9732} & 0.8200 \\
		\cline{2-9}
		& AP & \textbf{0.9846} & 0.8769 & 0.9642 & 0.6995 & 0.7328 & \textbf{0.9771} & 0.8344 \\
		\hline
		$|\cdot|_2$ & AUC & \textbf{0.9837} & 0.9005 & 0.9582 & 0.6599 & 0.7297 & \textbf{0.9754} & 0.8243 \\
		\cline{2-9}
		& AP & \textbf{0.9852} & 0.8782 & 0.9593 & 0.6962 & 0.7431 & \textbf{0.9791} & 0.8383 \\
		\hline
	\end{tabular}
\end{table*}
In this section, we evaluate the ability of node representations via link prediction which is a typical task of network representation learning. The purpose of link prediction task is to discover missing edges given a network with a certain fraction of edges removed.

Facebook network dataset is implemented in our link prediction experiments. To conduct the link prediction task, we first need to divide the original network into the training set and testing set: we randomly remove a certain portion of edges as positive sample in testing set, and the edges in the remaining network as positive sample in training set. To generate negative samples, we randomly sample an equal number of edges from the network which have no edge connecting them. After splitting the network we should ensure the remaining network is connected. Then we use network representation learning methods to obtain the node representation on the remaining connected networks. After training, we can obtain edge representations via some operators in Table \ref{tab:edge_feat} of two nodes representations and use it to predict missing edges as a binary classification by Logistic Regression. We report the results in terms of AUC score and average precision(AP) score\cite{Kipf2016Variational}. Table \ref{tab:lp15}, \ref{tab:lp30}, \ref{tab:lp50} show the result of removing $15\%$, $30\%$ and $50\%$ of edges in the Facebook network dataset. We bold the top two results for each row. It can be seen that our method performs well in most cases. And NEDP, DeepWalk and SDNE all achieve promising results in the link prediction task.

We can see that the same model are different from each other by using different operators to construct edge representations. However the model proposed in this paper ranks very high and the results are convincing in terms of the five constructive methods.  Although the network structure becomes more and more sparse with the removal of more and more edges, our proposed model is still competitive. For example, we removed $50\%$ edges in Table \ref{tab:lp50}, the AP value of our model can still reach $0.8303$ compared to Stru2Vec which is $0.8328$ under the same operator. Depwalk, Struc2Vec, SDNE and other methods show excellent performance in link prediction experiments, even exceed the model proposed in this paper in the case of constructing edge representations by L1 regularization and L2 regularization. From the above observation, we can see that the deep network embedding model captures the transfer behavior between nodes based on the similarity of social roles in the network to model the edges in the network. The representations obtained by NEDP can be generalized to the link prediction task smoothly.

Although the baseline methods are state-of-the-art and very effective, they cannot perform well in all the tasks and all the datasets. For example, DeepWalk achieves great results in the classification task; however it does not satisfy the clustering, visualization and network reconstruction tasks. DeepWalk is robust to most of the datasets in the classification task. Our method achieves slightly better results than the best baseline method. Nevertheless, we illustrate the promising performance of our method in all the tasks including clustering, classification, visualization and network reconstruction. Even for the link prediction task, our model shows competitive performance with the state-of-the-art.


\subsection{Parameter Sensitivity}
In this section, we will investigate the parameter sensitivity, w.r.t. Walks per node $\gamma$, Walk length $l$ and Expected representation dimension $d$, in order to guide us in selecting the optimal parameters. We conduct the node classification task on Cora with 5:5 train-test ratio and use the $Accuracy$ to illustrate the performance of our model. We examine each parameter by fixing the other two parameters.

\begin{figure*}[t]
	\centering
	\subfloat[Dimension]{\includegraphics[width=0.30\textwidth]{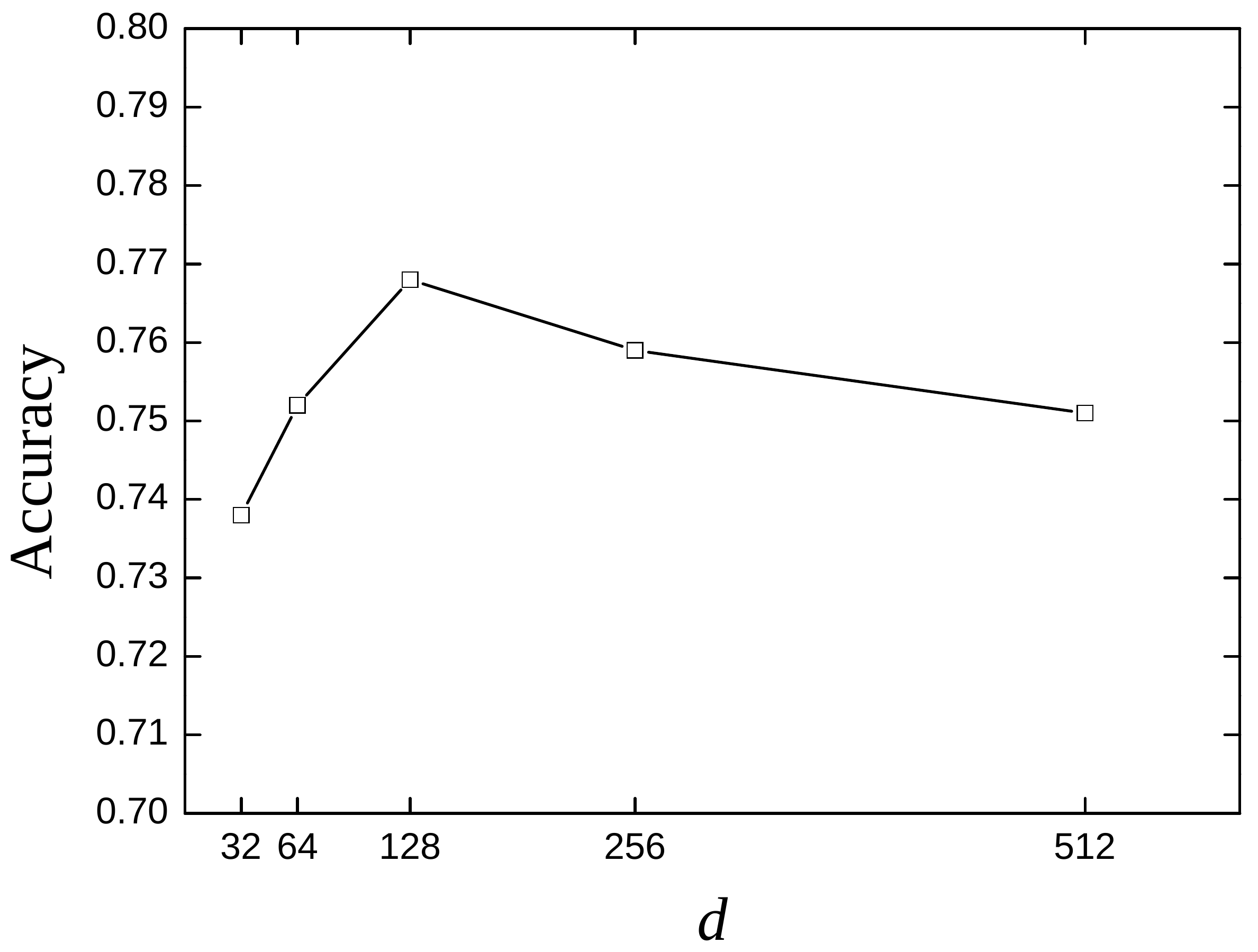}\label{fig:para_d}}
	\hspace{0.010cm}
	\subfloat[Walks per node]{\includegraphics[width=0.30\textwidth]{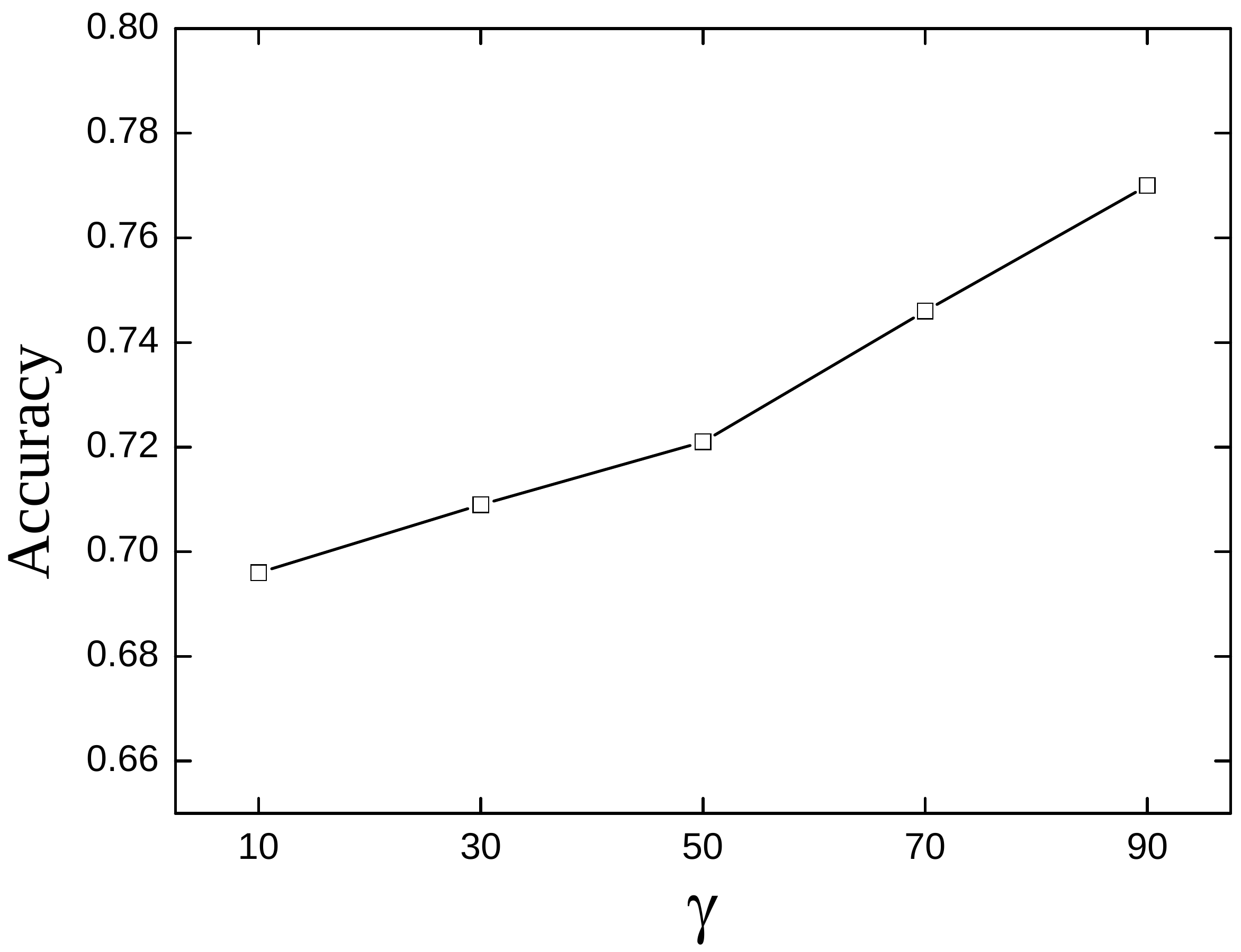}\label{fig:para_n}}
	\hspace{0.010cm}
	\subfloat[Walk length]{\includegraphics[width=0.30\textwidth]{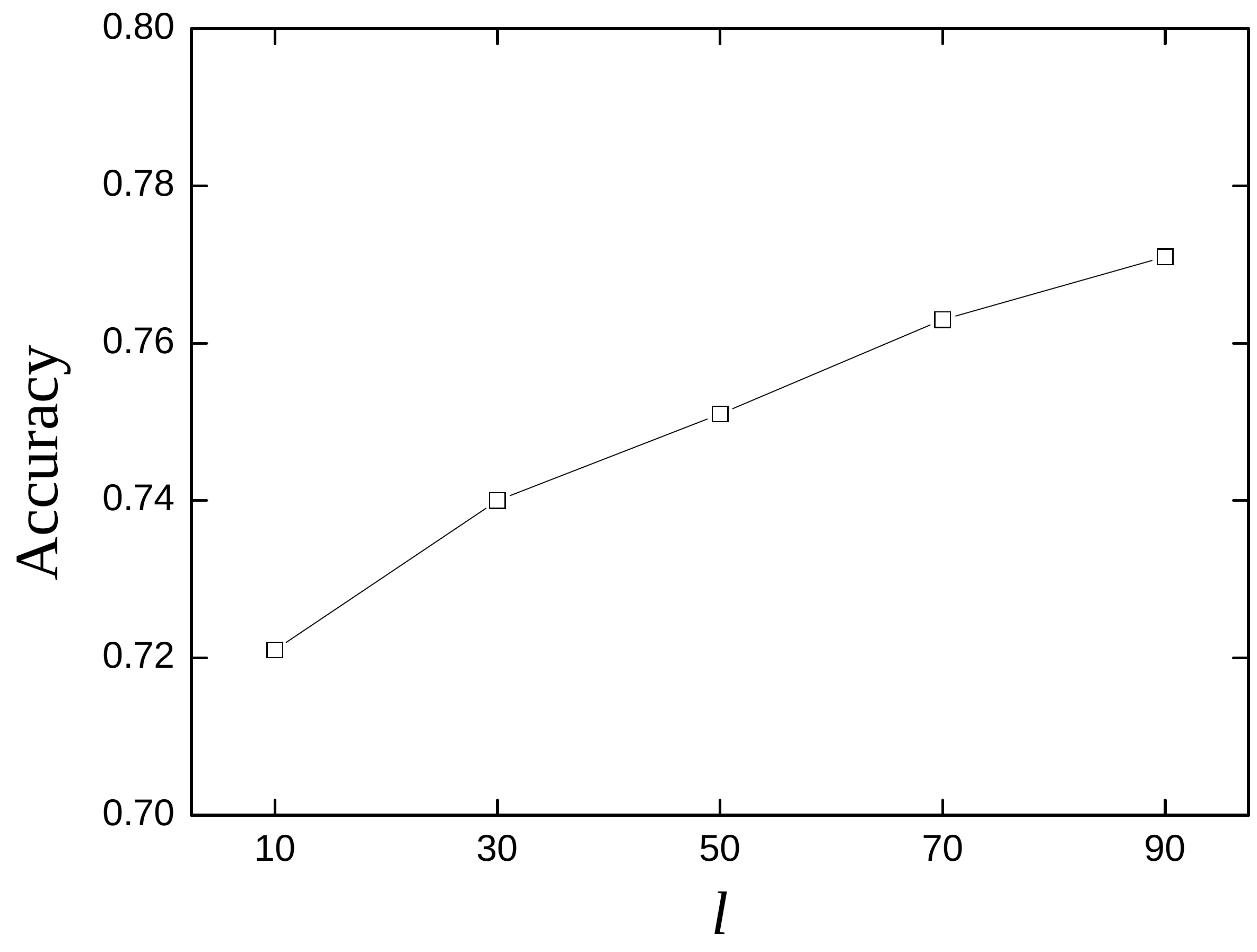}\label{fig:para_t}}
	\caption{ Parameter sensitivity analysis of our model on Cora with train ratio as 50\% }
	\label{fig:para}
\end{figure*}

Figure \ref{fig:para_d} shows the results about dimension $d$. As the dimension increases, the accuracy first rises rapidly and then slowly declines. The number 128 is the most appropriate dimension. Figure \ref{fig:para_n} reveals the effect of $\gamma$ and $l$ on model's performance. The accuracy score is positively related to $\gamma$ : as $\gamma$ increases, the accuracy score rises linearly. $\gamma$ corresponds to the capacity of the corpora, and the bigger $\gamma$ is, the richer the corpora is. The length $l$ of sequences generated by DW-Random Walk affects the global information of the node. LSTM can deal with long-term dependence sequence problems, so an increase in sequence length does not diminish the performance of the model.

\subsection{Performance w.r.t. Random Walks}
\begin{figure}
  \centering
  \includegraphics[scale=0.25]{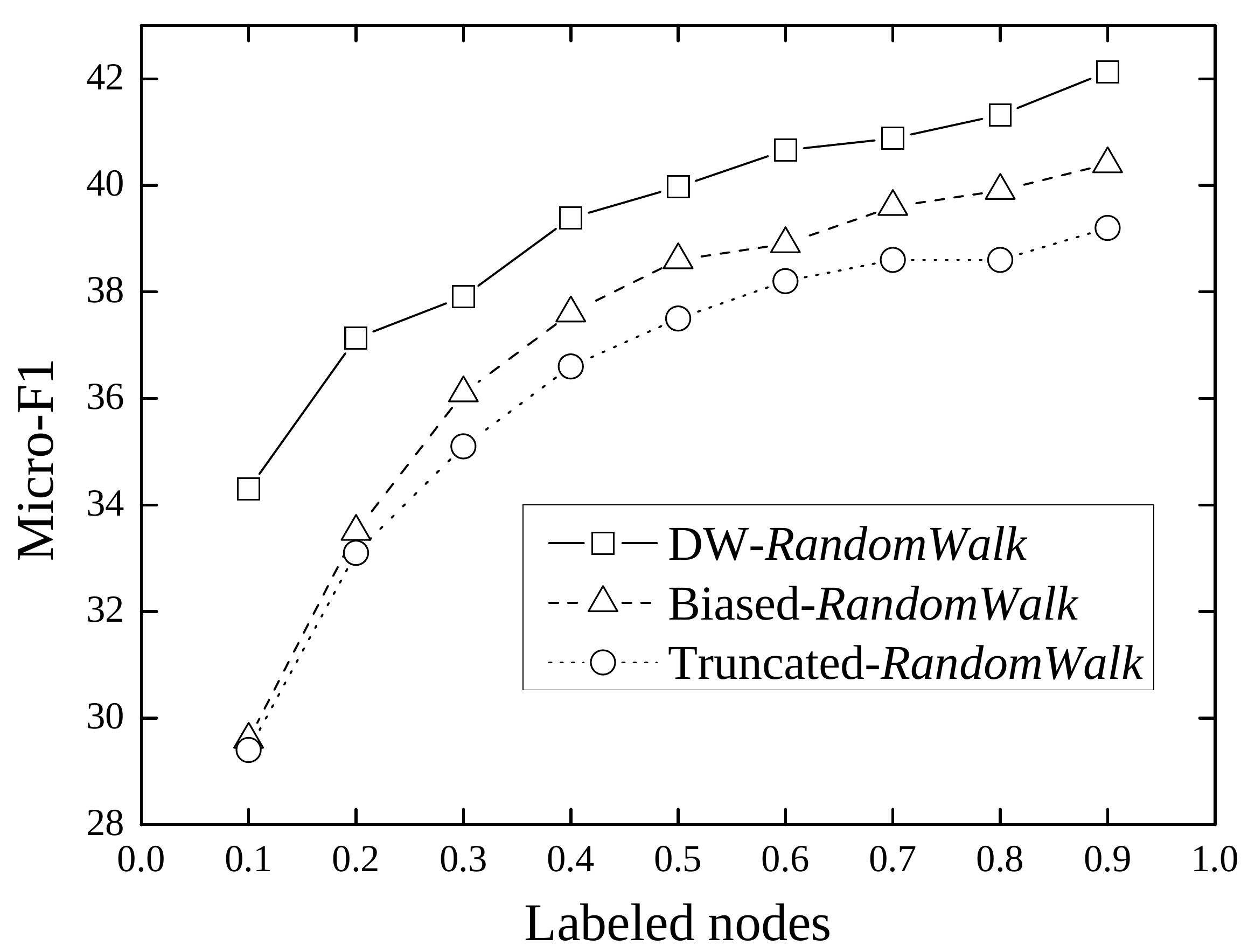}
  \caption{The results of different Random Walks}\label{fig:walk}
\end{figure}

In this section, we compare the performance of three types of Random Walks on the results, i.e., DW-$RandomWalk$, Truncated-$RandomWalk$, and Biased-$RandomWalk$. We sample the node sequences on the BlogCatalog dataset with the three different Random Walk methods, and then use the NEDP model to learn the representation. Figure \ref{fig:walk} shows the results. We can see that our DW-$RandomWalk$ method outperforms the others. Biased-$RandomWalk$ is lightly better than Truncated-$RandomWalk$. Especially in case the number of labeled nodes is small, the performance of DW-$RandomWalk$ method is significantly better than the others. In fact, when trained with only 20\% of the nodes labeled, DW-$RandomWalk$ performs better than Truncated-$RandomWalk$ and Biased-$RandomWalk$. It can be seen that considering the properties (such as weight and degree) of the network during random walks helps improve the performance of the model. DW-$RandomWalk$ utilizes both two important properties of the network, i.e., the degree of the node and weight of the edge, which allows exploring the functional role of nodes. The degree of the node can often reflect its importance in the network. By considering the degrees and weights jointly, we can traverse more diverse neighbor nodes. By introducing degree and weight information, we also avoid other hyper-parameters, such as $p$ and $q$ for the Biased-$RandomWalk$.

\section{Discussion and Conclusion}
Network structured data poses a great challenge for the traditional machine learning algorithms. Recent years researchers pay attention to designing feature vectors for learning algorithms. Structural function and neighborhood context are potential knowledge of the network structured data. And information transferring characterizes the structural roles of the nodes and connections, also the network evolution behaviors. This work introduces embedding layers to the traditional deep prediction model to learn the representation for each network node. To the best of our knowledge, it is the first time of employing a prediction model to generate new node representations.

Random walk shows its nice ability on capturing the global neighborhood context in the former researches, such as Deepwalk \cite{Perozzi2014} and Node2vec \cite{Grover2016}. This paper further suggests a degree-weight biased random walk model capture the functional and global network context information. In the network domain, the social role of the node is mostly dependent on the degree. And the weight denotes the relationships among  nodes. The new walk stage can guide the walking process to capture the major social role and global structural context. The prediction model is mainly used to embed the network structure properties. Besides, it can also preserve the transfer possibilities among the nodes. To capture the local network structure, we propose a Laplacian supervised embedding space optimization method following the embedding layer of the model. Laplacian Eigenmaps ensures that two connected nodes are close in low-dimensional space, so it can extract local information from the network. Our proposed NEDP method could generate a robust representation. 

Experiments on multiple datasets are conducted to evaluate the network representation generated by the method. We designed five experiments which prove that NEDP model is superior to other baselines in comprehensive aspects. By making full use of the network structure and deep learning technology, our model has stronger generalization ability.

Our future work will focus on how to learn representation in dynamic network and designing an end-to-end representation learning framework.


%



\section*{Acknowledgment}

This work is supported in part by the National Natural Science Foundation of China (No. 61971388, U1706218, 41576011). X. Sun is supported by Humboldt Research Fellowship Programme for Experienced Researchers. We would like to express our sincere appreciation to the anonymous reviewers.

\ifCLASSOPTIONcaptionsoff
  \newpage
\fi



\bibliographystyle{IEEEtran}
\bibliography{reference.bib}
%
%
%

%

\begin{IEEEbiography}[{\includegraphics[width=1in,height=1.25in,clip,keepaspectratio]{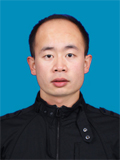}}]{Xin Sun}
is currently an associate professor at Ocean University of China, P.R.China. He received his Ph.D. degree from the College of Computer Science and Technology at Jilin University in 2013. He did the Post-Doc research (2016-2017) in the department of computer science at the Ludwig-Maximilians-Universit\"{a}t M\"{u}nchen, Germany. His current research interests include machine learning, data mining and computer vision.
\end{IEEEbiography}
\vspace{-20 mm}
\begin{IEEEbiography}[{\includegraphics[width=1in,height=1.25in,clip,keepaspectratio]{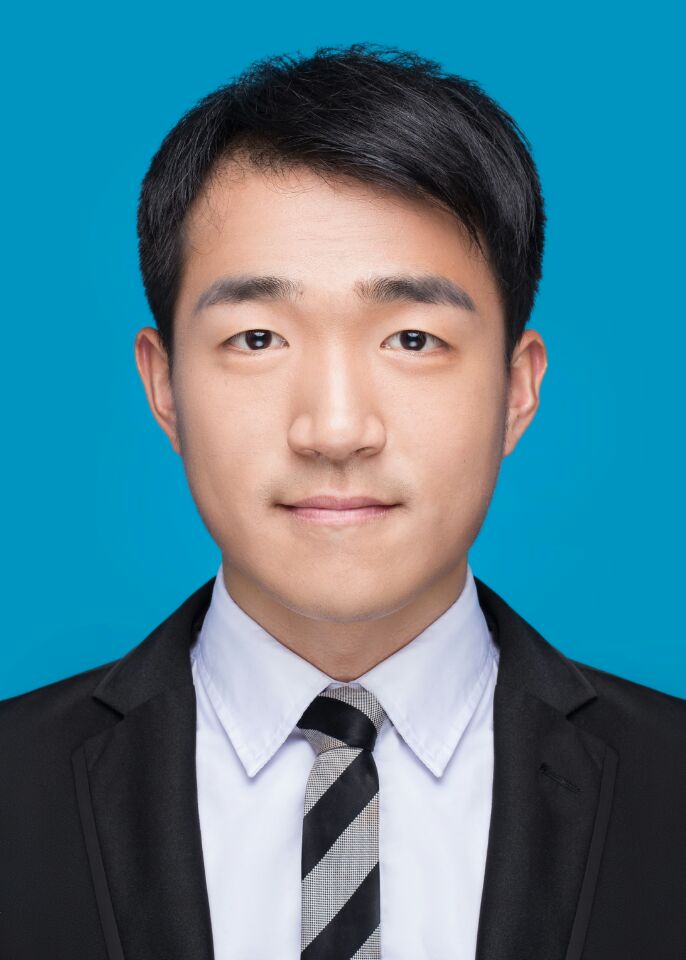}}]{Zenghui Song}
received the M.Sc. degree from the Department of Computer Science and Technology, Ocean University of China, P.R. China, in 2019. His research interests include data mining and deep learning.
\end{IEEEbiography}
\vspace{-20 mm}
\begin{IEEEbiography}[{\includegraphics[width=1in,height=1.25in,clip,keepaspectratio]{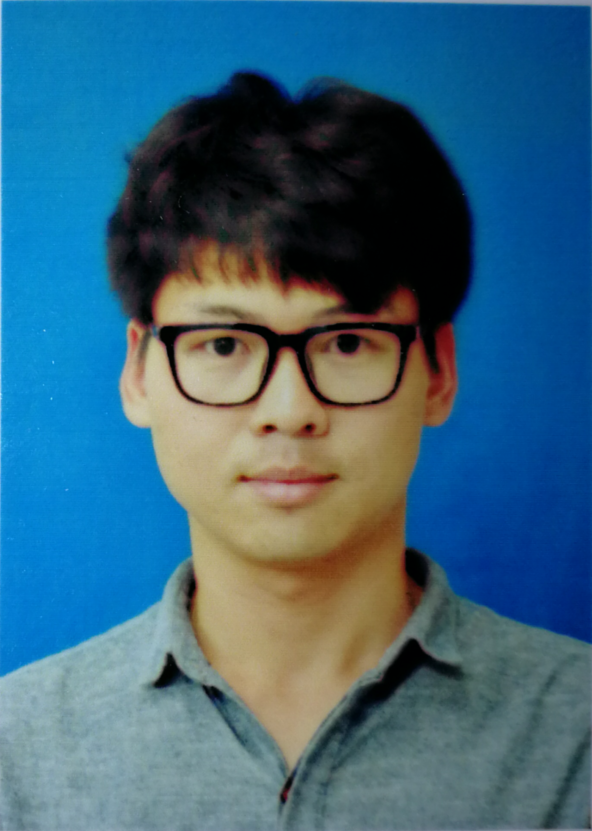}}]{Yongbo Yu}
received his B.Sc in the Department of Computer Science and Technology, Ocean University of China, in 2017. His research focuses on data mining for social network analysis.
\end{IEEEbiography}
\vspace{-20 mm}
\begin{IEEEbiography}[{\includegraphics[width=1in,height=1.25in,clip,keepaspectratio]{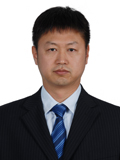}}]{Junyu Dong}
received the Ph.D. degree from the School of Mathematical and Computer Sciences, Heriot-Watt University, Edinburgh, U.K., in 2003. He is currently a professor and the Head of the Department of Computer Science and Technology, Ocean University of China. His current research interests include machine learning, video analysis, and image processing.
\end{IEEEbiography}
\vspace{-20 mm}
\begin{IEEEbiography}[{\includegraphics[width=1in,height=1.25in,clip,keepaspectratio]{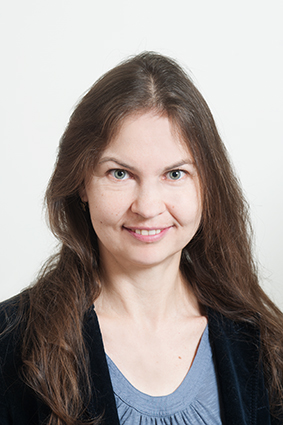}}]{Claudia Plant}
received her Ph.D. degree in 2007 and currently is full professor for Data Mining at University of Vienna. Her research focuses on database-related data mining, especially clustering, parameter-free data mining, and integrative mining of heterogeneous data. She not only published in top-level database and data mining conferences like KDD and ICDM but also in application-related toplevel journals. She received 3 awards including the ICDM best paper award 2014.
\end{IEEEbiography}
\vspace{-20 mm}
\begin{IEEEbiography}[{\includegraphics[width=1in,height=1.25in,clip,keepaspectratio]{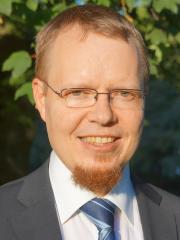}}]{Christian B\"{o}hm}
is professor of informatics at Ludwig-Maximilians-Universit\"{a}t M\"{u}nchen, Germany. He received his Ph.D. in 1998 and his habilitation in 2001. His former affiliations include the Technische Universitat M\"{u}nchen and UMIT Hall in Tirol, Austria. His research focus is data mining, particularly index structures and clustering algorithms. He has received 4 research awards including the SIGMOD best paper award 1997 and the SIAM SDM best paper honorable mention award 2008.
\end{IEEEbiography}
%
%
%





\end{document}